\documentclass[runningheads]{llncs}
\usepackage{graphicx}
\usepackage{tikz}
\usepackage{comment}
\usepackage{amsmath,amssymb} % define this before the line numbering.
\usepackage{color}

\usepackage[accsupp]{axessibility}  % Improves PDF readability for those with disabilities.

\usepackage{graphicx}
\usepackage{amsmath}
\usepackage{amssymb}
\usepackage{booktabs}

\usepackage{multirow}
\usepackage{caption}
\usepackage{subcaption}
\usepackage{array}
\usepackage{pbox}
\usepackage{wrapfig}
\usepackage{hyperref}
\begin{document}
% \linenumbers
\pagestyle{headings}
\mainmatter
\def\ECCVSubNumber{5424}  % Insert your submission number here

\title{Are Vision Transformers Robust to Patch Perturbations?} % Replace with your title

\begin{comment}
\titlerunning{ECCV-22 submission ID \ECCVSubNumber} 
\authorrunning{ECCV-22 submission ID \ECCVSubNumber} 
\author{Anonymous ECCV submission}
\institute{Paper ID \ECCVSubNumber}
\end{comment}
%******************

%\begin{comment}
\titlerunning{Are Vision Transformers Robust to Patch Perturbations?}
\author{Jindong Gu\inst{1} \and
Volker Tresp\inst{1}\and
Yao Qin\inst{2}}
\authorrunning{J. Gu et al.}
\institute{$^1 \,$University of Munich \hspace{0.2cm} $^2 \,$Google Research}
%\end{comment}
%******************
\maketitle

\begin{abstract}
Recent advances in Vision Transformer (ViT) have demonstrated its impressive performance in image classification, which makes it a promising alternative to Convolutional Neural Network (CNN). Unlike CNNs, ViT represents an input image as a sequence of image patches. The patch-based input image representation makes the following question interesting: How does ViT perform when individual input image patches are perturbed with natural corruptions or adversarial perturbations, compared to CNNs? In this work, we study the robustness of ViT to patch-wise perturbations. Surprisingly, we {find} that ViTs are more robust to naturally corrupted patches than CNNs, whereas they are more vulnerable to adversarial patches. Furthermore, we discover that the attention mechanism greatly affects the robustness of vision transformers. Specifically, the attention module can help improve the robustness of ViT by effectively ignoring natural corrupted patches. However, when ViTs are attacked by an adversary, the attention mechanism can be easily fooled to focus more on the adversarially perturbed patches and cause a mistake. Based on our analysis, we propose a simple temperature-scaling based method to {improve} the robustness of ViT against adversarial patches. Extensive qualitative and quantitative experiments are performed to support our findings, understanding, and improvement of ViT robustness to patch-wise perturbations across a set of transformer-based architectures.
\keywords{Understanding Vision Transformer, Adversarial Robustness}
\end{abstract}

\begin{figure}[!ht]
    \hspace{0.1cm}
    \begin{subfigure}[b]{0.17\textwidth}
    \centering
    \includegraphics[scale=0.2]{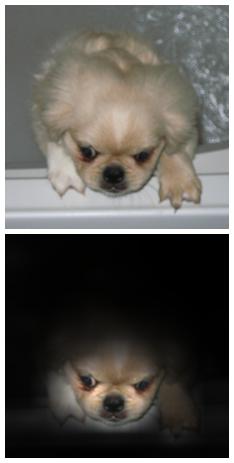}
    \caption{\scriptsize Clean Image}
    \label{fig:teaser_clean}
    \end{subfigure} \hspace{-0.15cm}
    \begin{subfigure}[b]{0.4\textwidth}
    \centering
    \includegraphics[scale=0.2]{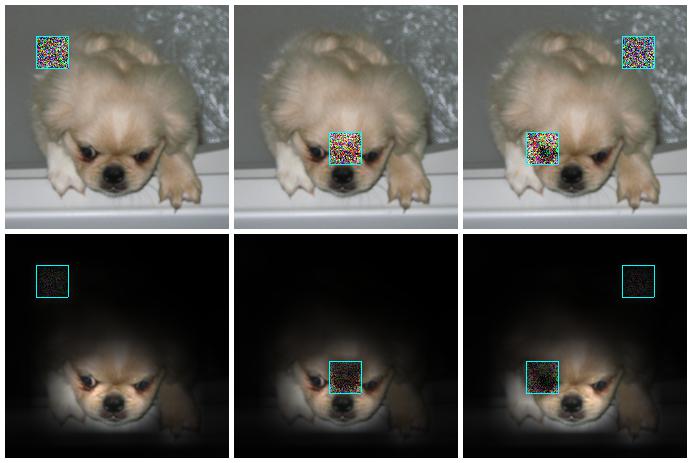}
    \caption{\scriptsize with Naturally Corrupted Patch}
    \label{fig:teaser_nat}
    \end{subfigure} \hspace{0.05cm}
    \begin{subfigure}[b]{0.4\textwidth}
    \centering
    \includegraphics[scale=0.2]{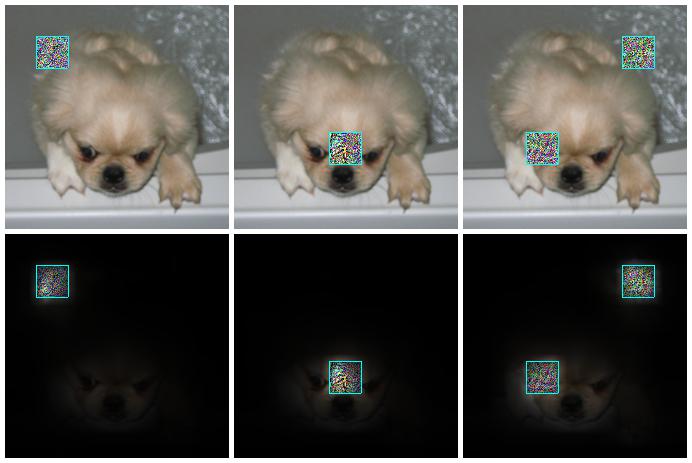}
    \caption{\scriptsize with Adversarial Patch}
    \label{fig:teaser_adv}
    \end{subfigure} \vspace{0.05cm}
    \caption{Images with patch-wise perturbations (top) and their corresponding attention maps (bottom). The attention mechanism in ViT can effectively ignore the naturally corrupted patches to maintain a correct prediction in Fig. b, whereas it is forced to focus on the adversarial patches to make a mistake in Fig. c. The images with corrupted patches (Fig. b) are all correctly classified. The images with adversary patches (Fig. c) are misclassified as \textit{dragonfly}, \textit{axolotl}, and \textit{lampshade}, respectively.}
    \label{fig:teaser}
\end{figure}

\vspace{-0.5cm}
\section{Introduction}
Recently, Vision Transformer (ViT) has demonstrated impressive performance ~\cite{dosovitskiy2020image,touvron2021training,wu2020visual,xiao2021early,graham2021levit,chen2021visformer,han2021transformer,chen2021crossvit,liu2021swin}, which makes it become a potential alternative to convolutional neural networks (CNNs). Meanwhile, the robustness of ViT has also received great attention~\cite{bhojanapalli2021understanding,joshi2021adversarial,qin2021understanding,salman2021certified,shao2021adversarial,shi2021decision,tang2021robustart}. On the one hand, it is important to improve its robustness for safe deployment in the real world. On the other hand, diagnosing the vulnerability of ViT can also give us a deeper understanding of its underlying working mechanisms. Existing works have intensively studied the robustness of ViT and CNNs when the whole input image is perturbed with natural corruptions or adversarial perturbations~\cite{bhojanapalli2021understanding,shao2021adversarial,mahmood2021robustness,bai2021transformers,aldahdooh2021reveal}. Unlike CNNs, ViT processes the input image as a sequence of image patches. Then, a self-attention mechanism is applied to aggregate information from all patches. Based on the special patch-based architecture of ViT, we mainly focus on studying the robustness of ViT to patch-wise perturbations. 

In this work, two typical types of perturbations are considered to compare the robustness between ViTs and CNN (e.g., ResNets~\cite{he2016deep}). One is natural corruptions~\cite{hendrycks2019benchmarking}, which is to test models' robustness under distributional shift. The other is adversarial perturbations~\cite{szegedy2013intriguing,goodfellow2014explaining}, which are created by an adversary to specifically fool a model to make a wrong prediction. Surprisingly, we find ViT does \emph{not always} perform more robustly than ResNet. When individual image patches are naturally corrupted, ViT is more robust compared to ResNet. However, when input image patch(s) are adversarially attacked, ViT shows a higher vulnerability than ResNet.

Digging down further, we revealed that ViT's stronger robustness to natural corrupted patches and higher vulnerability against adversarial patches are both caused by the attention mechanism. Specifically, the self-attention mechanism of ViT can effectively ignore the natural patch corruption, while it's also easy to manipulate the self-attention mechanism to focus on an adversarial patch. This is well supported by rollout attention visualization~\cite{abnar2020quantifying} on ViT. As shown in Fig.~\ref{fig:teaser} (a), ViT successfully attends to the class-relevant features on the clean image, \textit{i.e.}, the head of the dog. When one or more patches are perturbed with natural corruptions, shown in Fig.~\ref{fig:teaser} (b), ViT can effectively ignore the corrupted patches and still focus on the main foreground to make a correct prediction. In Fig.~\ref{fig:teaser} (b), the attention weights on the positions of naturally corrupted patches are much smaller even when the patches appear on the foreground. In contrast, when the patches are perturbed with adversarial perturbations by an adversary, ViT is successfully fooled to make a wrong prediction, as shown in Fig.~\ref{fig:teaser} (c). This is because the attention of ViT is misled to focus on the adversarial patch instead.

Based on this understanding that the attention mechanism leads to the vulnerability of ViT against adversarial patches, we propose a simple Smoothed Attention to discourage the attention mechanism to a single patch. Specifically, we use a temperature to smooth the attention weights computed by a \textit{softmax} operation in the attention. In this way, a single patch can hardly dominate patch embeddings in the next layer, which can effectively improve the robustness of ViT against adversarial patch attacks.

Our main contributions can be summarized as follows:
\begin{itemize}
  \item 
  \textbf{Finding}: Based on a fair comparison, we {discover} that ViT is more robust to natural patch corruption than ResNet, whereas it is more vulnerable to adversarial patch perturbation. 
  \item 
  \textbf{Understanding}: We reveal that the self-attention mechanism can effectively ignore natural corrupted patches to maintain a correct prediction but be easily fooled to focus on adversarial patches to make a mistake.
  \item 
  \textbf{Improvement}: Inspired by our understanding, we propose Smoothed Attention, which can effectively {improve} the robustness of ViT against adversarial patches by discouraging the attention to a single patch.
\end{itemize}

\section{Related Work}
\textbf{Robustness of Vision Transformer.}
The robustness of ViT have achieved great attention due to its great success~\cite{bhojanapalli2021understanding,naseer2021intriguing,shao2021adversarial,benz2021adversarial,mahmood2021robustness,bai2021transformers,naseer2021improving,aldahdooh2021reveal,salman2021certified,yu2021mia,hu2021inheritance,mao2021rethinking,mao2021towards,naseer2021intriguing,qin2021understanding}.
On the one hand, \cite{bhojanapalli2021understanding,paul2021vision} show that vision transformers are more robust to natural corruptions~\cite{hendrycks2019benchmarking} compared to CNNs. On the other hand, ~\cite{bhojanapalli2021understanding,shao2021adversarial,paul2021vision} demonstrate that ViT achieves higher adversarial robustness than CNNs under adversarial attacks. These existing works, however, mainly focus on investigating the robustness of ViT when a whole image is naturally corrupted or adversarially perturbed. Instead, our work focuses on patch perturbation, given the patch-based architecture trait of ViT. The patch-based attack~\cite{joshi2021adversarial,fu2021patch} and defense~\cite{mu2021defending,shi2021decision} methods have also been proposed recently. Different from their work, we aim to understand the robustness of patch-based architectures under patch-based natural corruption and adversarial patch perturbation.

\vspace{0.05cm}
\noindent\textbf{Adversarial Patch Attack.}
The seminal work~\cite{papernot2016limitations} shows that adversarial examples can be created by perturbing only a small amount of input pixels. Further, \cite{brown2017adversarial,liu2020bias} successfully creates universal, robust, and targeted adversarial patches. These adversarial patches therein are often placed on the main object in the images. The works~\cite{BMVC2016_137,metzen2021meta} shows that effective adversarial patches can be created without access to the target model. However, both universal patch attacks and black-box attacks are weak to be used for our study. They can only achieve very low fooling rates when a single patch of ViT (only 0.5\% of image) is attacked. In contrast, the white-box attack~\cite{karmon2018lavan,liu2019perceptual,wang2021universal,qian2020visually,luo2021generating} can fool models by attacking only a very small patch. In this work, we apply the most popular adversarial patch attack in \cite{karmon2018lavan} to both ViT and CNNs for our study.

\section{Experimental Settings to Compare ViT and ResNet}
\label{sec:exp_set}

\begin{table}[t]
\begin{center}
\footnotesize
\caption{Comparison of popular ResNet and ViT models. The difference in model robustness can not be blindly attributed to the model architectures. It can be caused by different training settings. WS, GN and WD correspond to Weight Standardization, Group Normalization and Weight Decay, respectively.}
\label{tab:sota_models}
\setlength\tabcolsep{0.16cm}
\begin{tabular}{l ccc ccc}
\toprule
Model  & Pretraining & DataAug &  Input Size &  WS & GN & WD\\
\midrule
ResNet~\cite{he2016deep}  & N  & N  &  224  & N  &  N  & Y \\
BiT~\cite{kolesnikov2020big}     & Y  & N  &  480  & Y  &  Y  & N \\
ViT~\cite{dosovitskiy2020image}      & Y  & N  &  224/384  & N  &  N  & N \\
DeiT~\cite{touvron2021training}    & N  & Y  &  224/384  & N  &  N  & N \\
\bottomrule
\end{tabular}
\end{center}
\vspace{-0.1cm}
\end{table}
%\vspace{2cm}

\noindent\textbf{Fair Base Models.} We list the state-of-the-art ResNet and ViT models and part of their training settings in Tab.~\ref{tab:sota_models}. The techniques applied to boost different models are different, \textit{e.g.}, pretraining. A recent work \cite{bai2021transformers} points out the necessity of a fair setting. Our investigation finds weight standardization and group normalization also have a significant impact on model robustness (More in Appendix A). This indicates that the difference in model robustness can not be blindly attributed to the model architectures if models are trained with different settings. Hence, we build fair models to compare ViT and ResNet as follows.

First, we follow~\cite{touvron2021training} to choose two pairs of fair model architectures, DeiT-small vs. ResNet50 and DeiT-tiny vs. ResNet18. The two models of each pair (\textit{i.e.} DeiT and its counter-part ResNet) are of similar model sizes. Further, we train ResNet50 and ResNet18 using the \textbf{exactly same setting} as DeiT-small and Deit-tiny in \cite{touvron2021training}. In this way, we make sure the two compared models, \textit{e.g.}, DeiT-samll and ResNet50, have similar model sizes, use the same training techniques, and achieve similar test accuracy (See Appendix A). The two fair base model pairs are used across this paper for a fair comparison.

\vspace{0.05cm}
\noindent\textbf{Adversarial Patch Attack.} We now introduce adversarial patch attack \cite{karmon2018lavan} used in our study. The first step is to specify a patch position and replace the original pixel values of the patch with random initialized noise $\delta$. The second step is to update the noise to minimize the probability of ground-truth class, \textit{i.e.} maximize the cross-entropy loss via multi-step gradient ascent \cite{madry2017towards}. The adversary patches are specified to align with input patches of DeiT.

\vspace{0.05cm}
\noindent\textbf{Evaluation Metric.} We use the standard metric \textbf{Fooling Rate (FR)} to evaluate the model robustness. First, we collect a set of images that are correctly classified by both models that we compare. The number of these collected images is denoted as $P$. When these images are perturbed with natural patch corruption or adversarial patch attack, we use $Q$ to denoted the number of images that are misclassified by the model. The Fooling Rate is then defined as FR = $\frac{Q}{P}$. The lower the FR is, the more robust the model is.

\begin{table*}[t]
\begin{center}
\caption{Fooling Rates (in \%) are reported. DeiT is more robust to naturally corrupted patches than ResNet, while it is significantly more vulnerable than ResNet against adversarial patches. Bold font is used to mark the lower fooling rate, which indicates the higher robustness.}
\label{tab:core_table}
\footnotesize
\setlength{\tabcolsep}{0.7em}
\begin{tabular}{c | p{0.75cm}p{0.75cm}p{0.75cm}p{0.75cm} | cccc}
\toprule
\multirow{2}{*}{Model}  &  \multicolumn{4}{c|}{\# Naturally Corrupted Patches} & \multicolumn{4}{c}{\# Adversarial Patches} \\
\cmidrule{2-9}
  & 32 & 96 & 160 & 196  & 1 & 2 & 3 & 4 \\
\midrule
ResNet50 & 3.7 & 18.2  & 43.4  & 49.8 & \textbf{30.6} & \textbf{59.3} & \textbf{77.1} & \textbf{87.2} \\
DeiT-small & \textbf{1.8} & \textbf{7.4} & \textbf{22.1} & \textbf{38.9} &  61.5 & 95.4 & 99.9 & 100 \\
\midrule
ResNet18 & 6.8 & 31.6 & 56.4 & 61.3 &\textbf{39.4} & \textbf{73.8} & \textbf{90.0} & \textbf{96.1} \\
DeiT-tiny & \textbf{6.4} & \textbf{14.6} & \textbf{35.8} & \textbf{55.9} & 63.3 & 95.8 & 99.9 & 100  \\
\bottomrule
\end{tabular}
\end{center}
\end{table*}

\section{ViT Robustness to Patch-wise Perturbations}
Following the setting in \cite{touvron2021training}, we train the models DeiT-small, ResNet50, DeiT-tiny, and ResNet18 on ImageNet 1k training data respectively. Note that no distillation is applied. The input size for training is $H=W=224$, and the patch size is set to $16$. Namely, there are 196 image patches totally in each image. We report the clean accuracy in Appendix A where DeiT and its counter-part ResNet show similar accuracy on clean images.

\subsection{Patch-wise Natural Corruption} 
First, we investigate the robustness of DeiT and ResNet to patch-based natural corruptions. Specifically, we randomly select 10k test images from ImageNet-1k validation dataset~\cite{deng2009imagenet} that are correctly classified by both DeiT and ResNet. Then for each image, we randomly sample $n$ input image patches $\boldsymbol{x}_i$ from  196 patches and perturb them with natural corruptions. 
As in \cite{hendrycks2019benchmarking}, 15 types of natural corruptions with the highest level are applied to the selected patches, respectively. The fooling rate of the patch-based natural corruption is computed over all the test images and all corruption types. We test DeiT and ResNet with the same naturally corrupted images for a fair comparison.

We find that both DeiT and ResNet hardly degrade their performance when a small number of patches are corrupted (\textit{e.g.}, 4). When we increase the number of patches, the difference between two architectures emerges: DeiT achieves a lower FR compared to its counter-part ResNet (See Tab.~\ref{tab:core_table}). This indicates that DeiT is more robust against naturally corrupted patches than ResNet. The same conclusion holds under the extreme case when the number of patches $n=196$. That is: the whole image is perturbed with natural corruptions. This is aligned with the observation in the existing work~\cite{bhojanapalli2021understanding} that vision transformers are more robust to ResNet under distributional shifts. More details on different corruption types are in Appendix B.

\begin{figure*}[t]
    \centering
    \begin{subfigure}[b]{0.45\textwidth}
        \includegraphics[width=\textwidth]{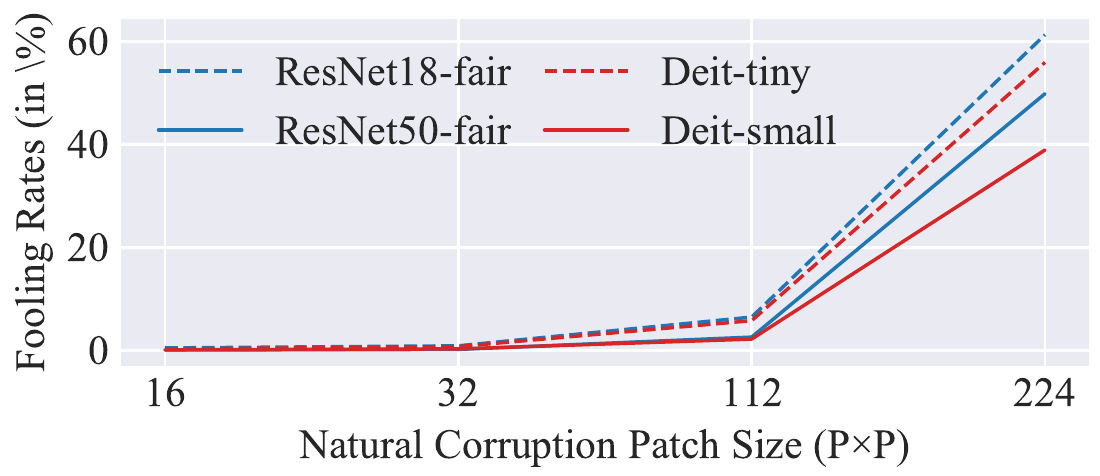}
    \label{fig:patch_size_nat}
    \end{subfigure} \hspace{0.2cm}
    \begin{subfigure}[b]{0.45\textwidth}
     \includegraphics[width=\textwidth]{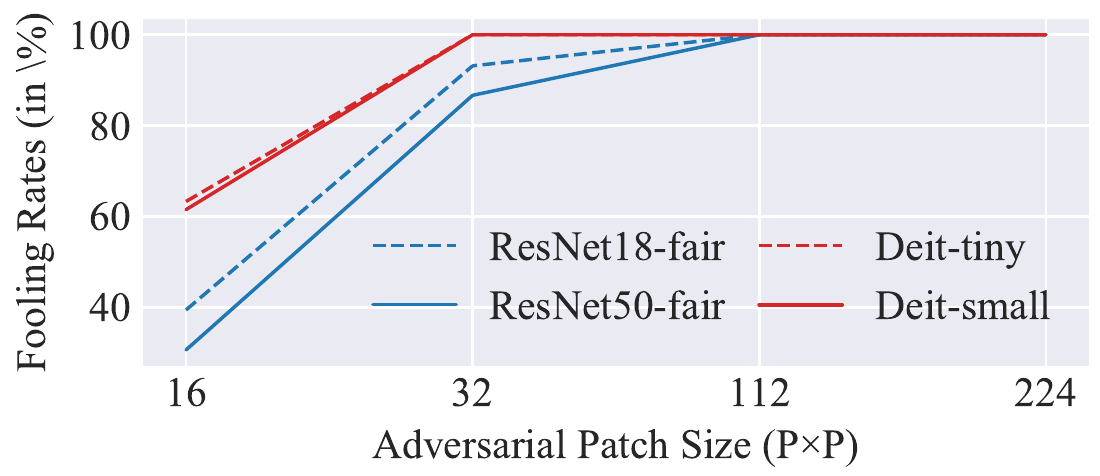}
      \label{fig:patch_size_adv}
    \end{subfigure}
    \vspace{-0.5cm}
    \caption{DeiT with red lines shows a smaller FR to natural patch corruption and a larger FR to adversarial patch of different sizes than counter-part ResNet.}
    \label{fig:patch_size}
\end{figure*}

In addition, we also increase the patch size of the perturbed patches, \textit{e.g.}, if the patch size of the corrupted patch is $32 \times 32$, it means that it covers 4 continuous and independent input patches as the input patch size is $16\times 16$. As shown in Fig.~\ref{fig:patch_size} (Left), even when the patch size of the perturbed patches becomes larger, DeiT (marked with red lines) is still more robust than its counter-part ResNet (marked with blue lines) to natural patch corruption.

\subsection{Patch-wise Adversarial Attack} In this section, we follow~\cite{karmon2018lavan} to generate adversarial patch attack and then compare the robustness of DeiT and ResNet against adversarial patch attack. We first randomly select the images that are correctly classified by both models from imagenet-1k validation daset. Following~\cite{karmon2018lavan}, the $\ell_{\infty}$-norm bound, the step size, and the attack iterations are set to 255/255, 2/255, and 10K respectively. Each reported FR score is averaged over 19.6k images.

As shown in Tab.~\ref{tab:core_table}, DeiT achieves much higher fooling rate than ResNet when one of the input image patches is perturbed with adversarial perturbation. This consistently holds even when we increase the number of adversarial patches, sufficiently supports that DeiT is more vunerable than ResNet against patch-wise adversarial perturbation.
When more than 4 patches ($\sim$2\% area of the input image) are attacked, both DeiT and ResNet can be successfully fooled with almost 100$\%$ FR.

When we attack a large continuous area of the input image by increasing the patch size of adversarial patches, the FR on DeiT is still much larger than counter-part ResNet until both models are fully fooled with 100$\%$ fooling rate. As shown in Fig.~\ref{fig:patch_size} (Right), DeiT (marked with red lines) consistently has higher FR than ResNet under different adversarial patch sizes. 

Taking above results together, we discover that DeiT is more robust to natural patch corruption than ResNet, whereas it is significantly more vulnerable to adversarial patch perturbation.

\begin{figure*}[!ht]
    \centering
    \begin{subfigure}[b]{0.99\textwidth}
        \includegraphics[width=0.96\textwidth]{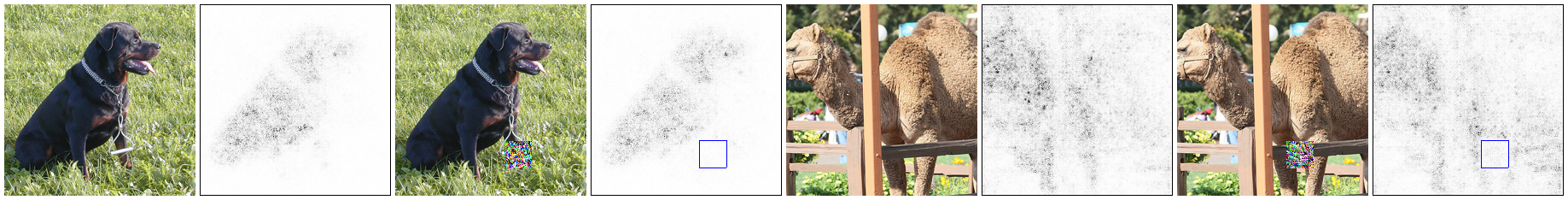}
        \includegraphics[scale=0.10]{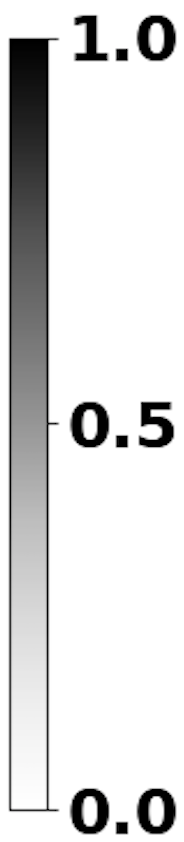}
    \caption{on ResNet50 under Adversary Patch Attack}
    \label{grad_vis_resnet50}
    \end{subfigure}
    
    \begin{subfigure}[b]{0.99\textwidth}
        \includegraphics[width=0.96\textwidth]{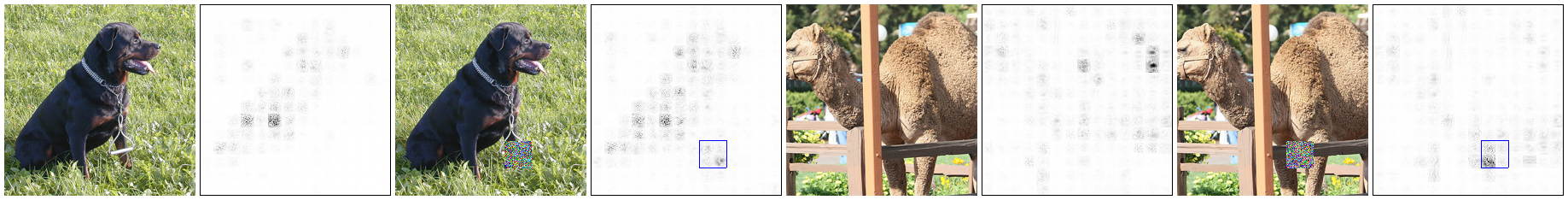}
        \includegraphics[scale=0.10]{figures/gray_bar.png}
        \caption{on DeiT-small under Adversary Patch Attack}
        \label{grad_vis_deit_small}
    \end{subfigure}
    \caption{Gradient Visualization. the clean image, the images with adversarial patches, and their corresponding gradient maps are visualized. We use a blue box on the gradient map to mark the location of the adversarial patch. The adversary patch on DeiT attracts attention, while the one on ResNet hardly do.}
    \label{fig:gradient_vis}
\end{figure*}

\section{Understanding ViT Robustness to Patch Perturbation}
\label{sec:attn_manip}
In this section, we design and conduct experiments to analyze the robustness of ViT. Especially, we aim to obtain deep understanding of how ViT performs when its input patches are perturbed with natural corruption or adversary patches.

\subsection{How ViT Attention Changes under Patch Perturbation?}
We visualize and analyze models' attention to understand the different robustness performance of DeiT and ResNet against patch-wise perturbations. Although there are many existing methods, \textit{e.g.}, \cite{selvaraju2017grad,shrikumar2017learning,zhou2016learning}, designed for CNNs to generate saliency maps, it is not clear yet how suitable to generalize them to vision transformers. Therefore, we follow \cite{karmon2018lavan} to choose the \textbf{model-agnostic} vanilla gradient visualization method to compare the gradient (saliency) map \cite{zeiler2014visualizing} of DeiT and ResNet. Specifically, we consider the case where DeiT and ResNet are attacked by adversarial patches. The gradient map is created as follow: we obtain the gradients of input examples towards the predicted classes, sum the absolute values of the gradients over three input channels, and visualize them by mapping the values into gray-scale saliency maps.

\textbf{Qualitative Evaluation.} As shown in Fig.~\ref{fig:gradient_vis} (a), when we use adversarial patch to attack a ResNet model, the gradient maps of the original images and the images with adversarial patch are similar. The observation is consistent with the one made in the previous work \cite{karmon2018lavan}. In contrast to the observation on ResNet, the adversarial patch can change the gradient map of DeiT by attracting more attention. As shown in Figure~\ref{fig:gradient_vis} (b), even though the main attention of DeiT is still on the object, part of the attention is misled to the adversarial patch. More visualizations are in Appendix C.

\textbf{Quantitative Evaluation.} We also measure our observation on the attention changes with the metrics in \cite{karmon2018lavan}. In each gradient map, we score each patch according to (1) the maximum absolute value within the patch (MAX); and (2) the sum of the absolute values within the patch (SUM). We first report the percentage of patches where the MAX is also the maximum of the whole gradient map. Then, we divide the SUM of the patch by the SUM of the all gradient values and report the percentage.

As reported in Tab.~\ref{tab:quan_exp}, the pixel with the maximum gradient value is more likely to fall inside the adversarial patch on DeiT, compared to that on ResNet. Similar behaviors can be observed in the metric of SUM. The quantitative experiment also supports our claims above that adversarial patches mislead DeiT by attracting more attention.

\begin{table*}[!t]
\begin{center}
\caption{Quantitative Evaluation. Each cell lists the percent of patches in which the maximum gradient value inside the patches is also the maximum of whole gradient map. SUM corresponds to the sum of element values inside patch divided by the sum of values in the whole gradient map. The average over all patches is reported.}
\label{tab:quan_exp}
\footnotesize
\setlength\tabcolsep{0.28cm}
\begin{tabular}{c  | cc | cc | cc | cc}
\toprule
  & \multicolumn{4}{c|}{Towards ground-truth Class} &  \multicolumn{4}{c}{Towards misclassified Class} \\
  \cmidrule{2-9}
  & \multicolumn{2}{c|}{SUM}  & \multicolumn{2}{c|}{MAX}  & \multicolumn{2}{c|}{SUM} & \multicolumn{2}{c}{MAX} \\
  \midrule
Patch Size & 16 & 32 &  16 &32 &  16&32 & 16&32 \\
\midrule
ResNet50 & 0.42& 1.40 & 0.17& 0.26 & 0.55 & 2.08 & 0.25 & 0.61 \\
DeiT-small & \textbf{1.98} & \textbf{5.33} & \textbf{8.3} & \textbf{8.39} & \textbf{2.21} & \textbf{6.31}  & \textbf{9.63} & \textbf{12.53} \\
\midrule 
ResNet18 & 0.24 & 0.74 & 0.01 &  0.02 &0.38 & 1.31  & 0.05& 0.13 \\
DeiT-tiny & \textbf{1.04} & \textbf{3.97} & \textbf{3.67} & \textbf{5.90} & \textbf{1.33} & \textbf{4.97} & \textbf{6.49} & \textbf{10.16} \\
\bottomrule
\end{tabular}
\end{center}
\end{table*}

\begin{figure*}[!t]
    \centering
    \begin{subfigure}[b]{0.96\textwidth}
        \includegraphics[width=\textwidth]{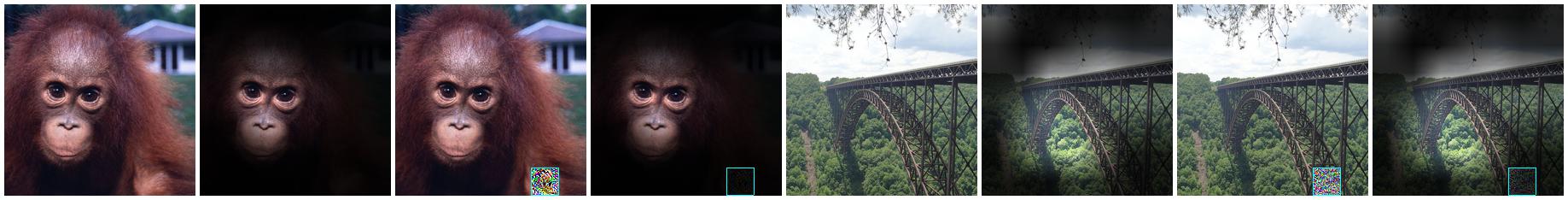}
    \caption{Attention on ResNet18 under Adversary Patch Attack}
    \label{grad_vis_resnet50}
    \end{subfigure}
    \begin{subfigure}[b]{0.96\textwidth}
        \includegraphics[width=\textwidth]{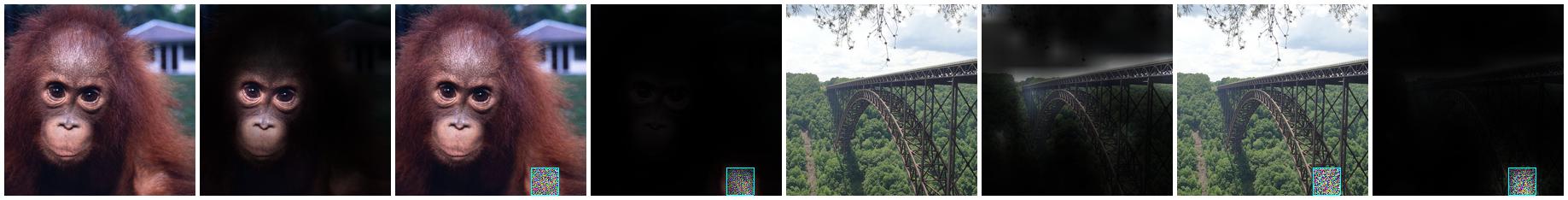}
        \caption{Attention on DeiT-tiny under Adversary Patch Attack}
        \label{grad_vis_deit_small}
    \end{subfigure}
    \caption{Attention Comparison between ResNet and DeiT under Patch Attack. The clean image, the adversarial images, and their corresponding attention are visualized. The adversary patch on DeiT attract attention, while the ones on ResNet hardly do.}
    \label{fig:rollout_feat_vis}
\end{figure*}

Besides the gradient analysis, another popular tool used to visualize ViT is Attention Rollout \cite{abnar2020quantifying}. To further confirm our claims above, we also visualize DeiT with Attention Rollout in Fig.~\ref{fig:rollout_feat_vis}. The rollout attention also shows that the attention of DeiT is attracted by adversarial patches. The attention rollout is not applicable to ResNet. As an extra check, we visualize and compare the feature maps of classifications on ResNet. The average of feature maps along the channel dimension is visualized as a mask on the original image. The visualization also supports the claims above. More visualizations are in Appendix D. Both qualitative and quantitative analysis verifies our claims that the adversarial patch can mislead the attention of DeiT by attactting it.

However, the gradient analysis is not available to compare ViT and ResNet on images with natural corrupted patches. When a small number of patch of input images are corrupted, both Deit and ResNet are still able to classify them correctly. The slight changes are not reflected in vanilla gradients since they are noisy. When a large area of the input image is corrupted, the gradient is very noisy and semantically not meaningful. Due to the lack of a fair visualization tool to compare DeiT and ResNet on naturally corrupted images, we apply Attention Rollout to DeiT and Feature Map Attention visualization to ResNet for comparing the their attention.

\begin{figure*}[!t]
    \centering
    \begin{subfigure}[b]{0.96\textwidth}
        \includegraphics[width=\textwidth]{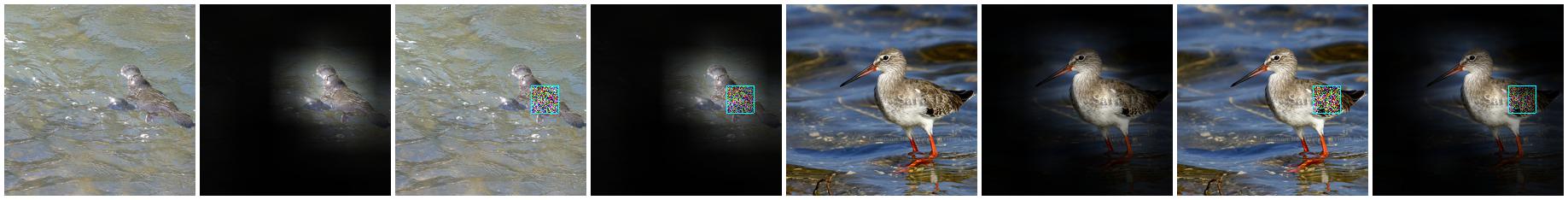}
        \caption{Attention on ResNet18 under Natural Patch Corruption}
    \label{grad_vis_resnet50}
    \end{subfigure}
    \begin{subfigure}[b]{0.96\textwidth}
        \includegraphics[width=\textwidth]{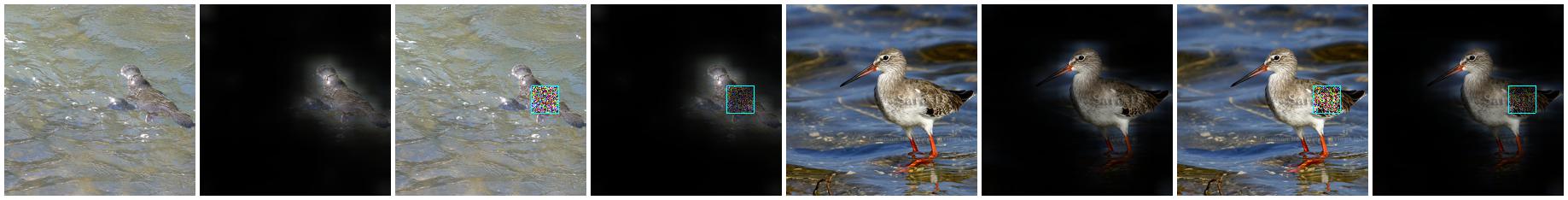}
        \caption{Attention on DeiT-tiny under Natural Patch Corruption}
        \label{grad_vis_deit_small}
    \end{subfigure}
    \caption{Attention Comparison between ResNet and DeiT under Natural Patch Corruption. The clean image, the naturally corrupted images, and their corresponding attention are visualized. The patch corruptions on DeiT are ignored by attending less to the corrupted patches, while the ones on ResNet are treated as normal patches.}
    \label{fig:rollout_feat_vis_nat}
\end{figure*}

The attention visualization of these images is shown in Fig.~\ref{fig:rollout_feat_vis_nat}. We can observe that ResNet treats the naturally corrupted patches as normal ones. The attention of ResNet on natually patch-corrupted images is almost the same as that on the clean ones. Unlike CNNs, DeiT attends less to the corrupted patches when they cover the main object. When the corrupted patches are placed in the background, the main attention of DeiT is still kept on the main object. More figures are in Appendix E.

\subsection{How Sensitive Is ViT Vulnerability to Attack Patch Positions?}
\begin{figure*}[!t]
    \centering
    \begin{subfigure}[b]{0.44\textwidth}
        \includegraphics[width=0.83\textwidth]{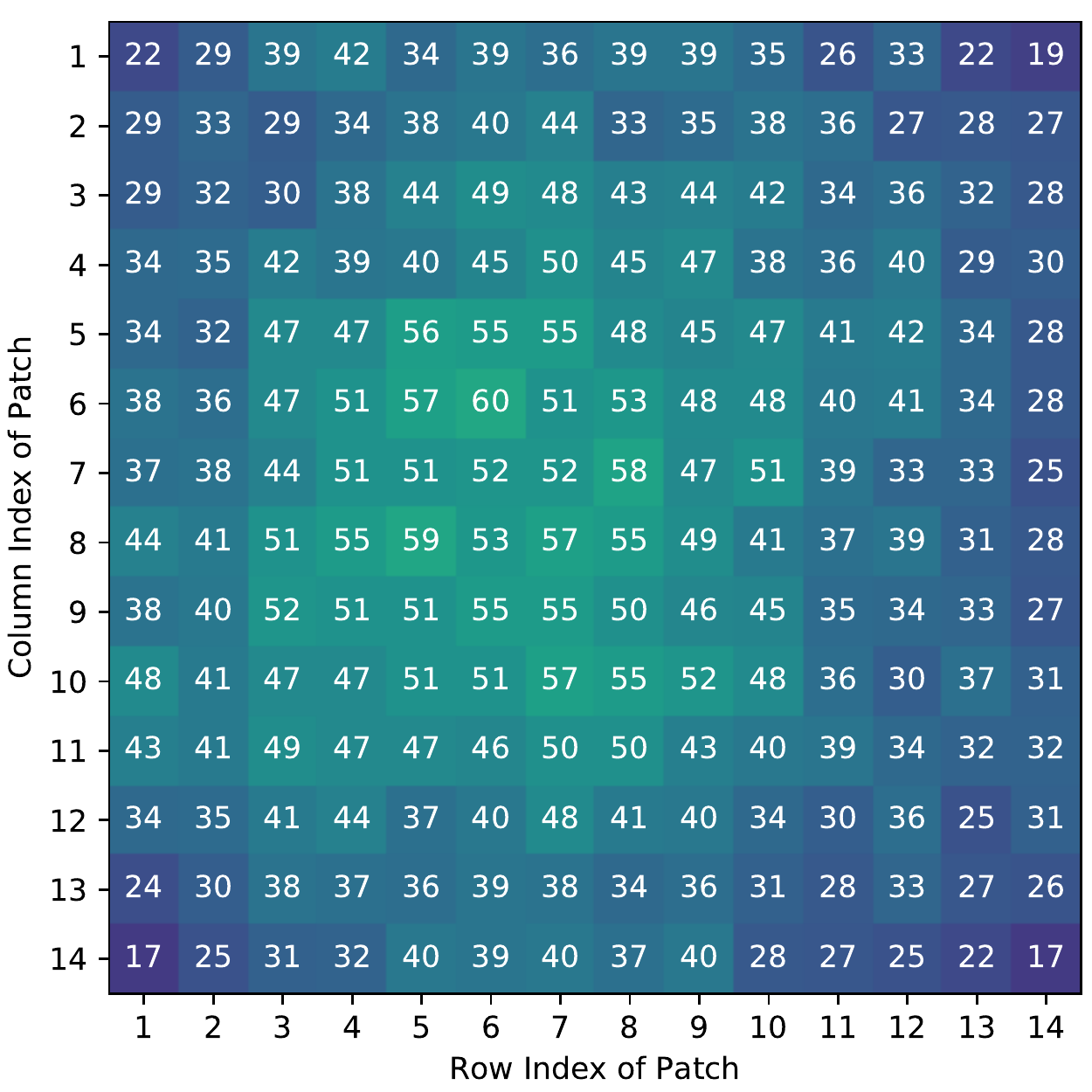}
    \caption{Patch Attack FRs on ResNet18}
    \end{subfigure} \hspace{0.2cm}
    \begin{subfigure}[b]{0.44\textwidth}
        \includegraphics[width=0.83\textwidth]{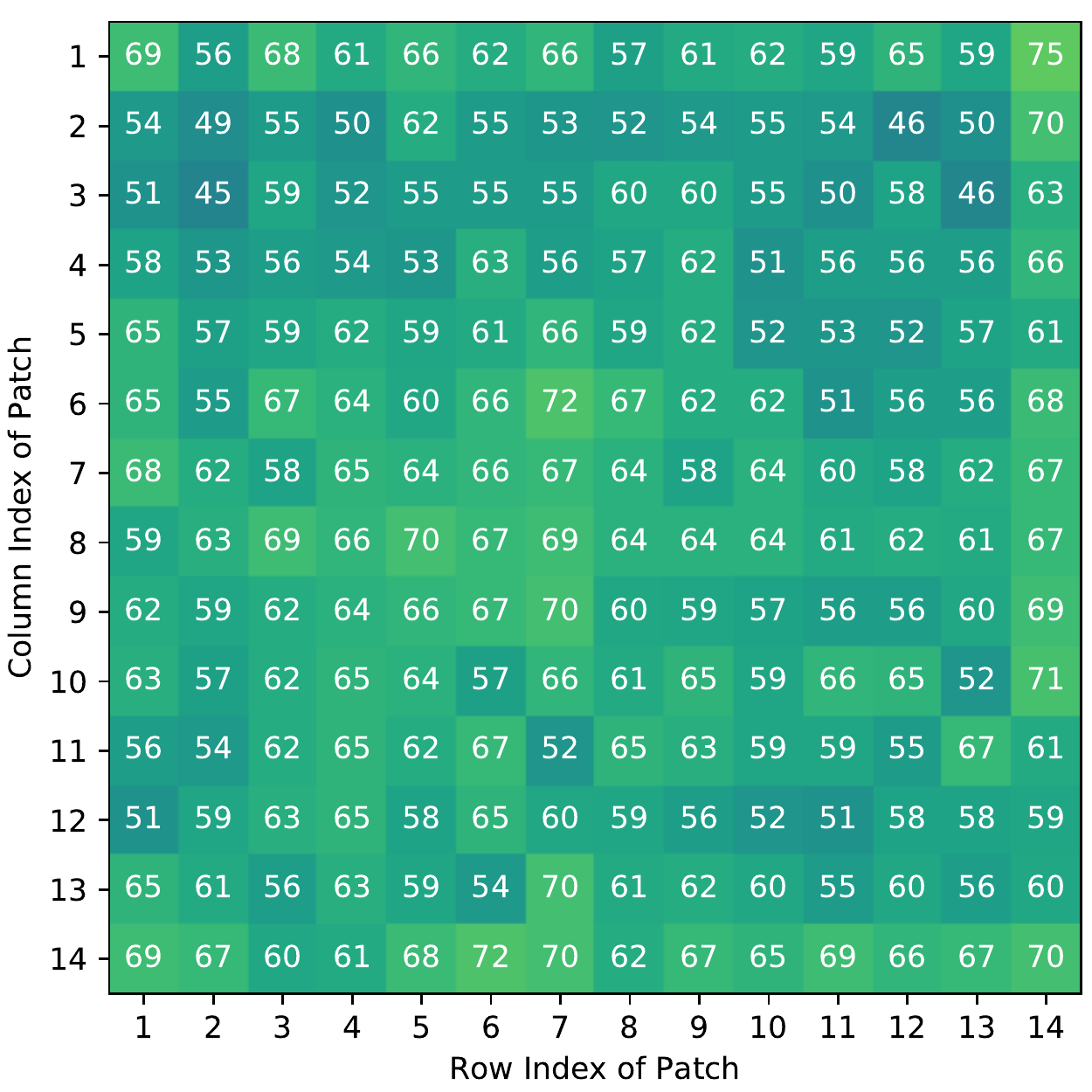}
        \caption{Patch Attack FRs on DeiT-tiny}
    \end{subfigure}
    \caption{Patch Attack FR (in \%) in each patch position is visualized. FRs in different patch positions of DeiT-tiny are similar, while the ones in ResNet18 are center-clustered.}
    \label{fig:FR_pattern_tiny_res18}
\end{figure*}

To investigate the sensitivity against the location of adversarial patch, we visualize the FR on each patch position in Fig.~\ref{fig:FR_pattern_tiny_res18}. We can clearly see that adversarial patch achieves higher FR when attacking DeiT-tiny than ResNet18 in different patch positions. Interestingly, we find that the FRs in different patch positions of DeiT-tiny are similar, while the ones in ResNet18 are center-clustered. A similar pattern is also found on DeiT-small and ResNet50 in Appendix F.

\begin{figure*}[!t]
    \centering
    \begin{subfigure}[b]{0.8\textwidth}
        \includegraphics[width=\textwidth]{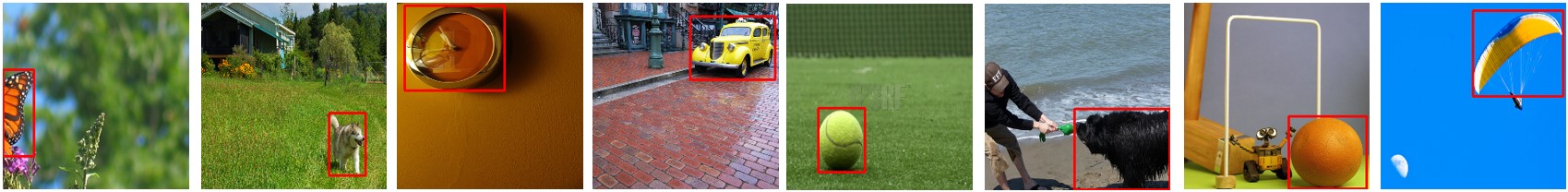}
    \caption{Corner-biased Images}
    \label{subfig:corner_bias}
    \end{subfigure}
    
    \begin{subfigure}[b]{0.8\textwidth}
        \includegraphics[width=\textwidth]{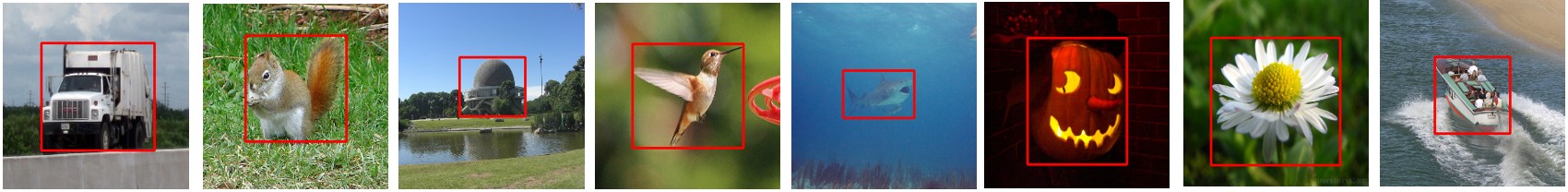}
        \caption{Center-biased Images}
        \label{subfig:center_bias}
    \end{subfigure}
    \caption{Collection of two sets of biased data. The fist set contains only images with corner-biased object(s), and the other set contains center-biased images.}
    \label{fig:bias_data}
\end{figure*}

Considering that ImageNet are center-biased where the main objects are often in the center of the images, we cannot attribute the different patterns to the model architecture difference without further investigation. 

Hence, we design the following experiments to disentangle the two factors, \textit{i.e.}, model architecture and data bias. Specifically, we select two sets of correctly classified images from ImageNet 1K validation dataset. As shown in Fig.~\ref{subfig:corner_bias}, the first set contains images with corner bias where the main object(s) is in the image corners. In contrast, the second set is more center-biased where the main object(s) is exactly in the central areas, as shown in Fig.~\ref{subfig:center_bias}.

We apply patch attack to corner-biased images (\textit{i.e.}, the first set) on ResNet. The FRs of patches in the center area are still significantly higher than the ones in the corner (See Appendix G). Based on this, we can conclude that such a relation of FRs to patch position on ResNet is caused by ResNet architectures instead of data bias. The reason behind this might be that pixels in the center can affect more neurons of ResNet than the ones in corners.

Similarly, we also apply patch attack to center-biased images (the second set) on DeiT. We observe that the FRs of all patch positions are still similar even the input data are highly center-biased (See Appendix H). Hence, we draw the conclusion that DeiT shows similar sensitivity to different input patches regardless of the content of the image. We conjecture it can be explained by the architecture trait of ViT, in which each patch equally interact with other patches regardless of its position.

\begin{table*}[!t]
\begin{center}
\caption{Transferability of adversarial patch across different patch positions of the the image. Translation X/Y stands for the number of pixels shifted in rows or columns. When they are shifted to cover other patches exactly, adversarial patches transfer well, otherwise not.}
\label{tab:trans}
\footnotesize
\setlength\tabcolsep{0.13cm}
\begin{tabular}{c| ccc | ccc | cc}
\toprule
Trans-(X,Y) & (0, 1) & \textbf{(0, 16)} & \textbf{(0, 32)} & (1, 0) & \textbf{(16, 0)} & \textbf{(32, 0)} & (1, 1) & \textbf{(16, 16)} \\
\midrule
ResNet50 & 0.06  & 0.31 & 0.48 & 0.06 & 0.18 & 0.40 & 0.08 & 0.35 \\
DeiT-small & 0.27  & \textbf{8.43} & \textbf{4.26} & 0.28 & \textbf{8.13} & \textbf{3.88} & 0.21 & \textbf{4.97} \\
\midrule
ResNet18  & 0.22 & 0.46 & 0.56  & 0.19 & 0.49 & 0.68 & 0.15 & 0.49  \\
DeiT-tiny & 2.54 & \textbf{29.15} & \textbf{18.19} & 2.30 & \textbf{28.37} & \textbf{17.32} & 2.11 & \textbf{21.23}  \\
\bottomrule
\end{tabular}
\end{center}
\vspace{-0.2cm}
\end{table*}

\subsection{Are Adversarial Patches on ViT Still Effective When Shifted?}
The work~\cite{karmon2018lavan} shows that the adversarial patch created on an image on ResNet is not effective anymore even if a single pixel is shifted away. Similarly, we also find that the adversarial patch perturbation on DeiT does not transfer as well when shifting a single pixel away. However, when an adversarial patch is shifted to exactly match another input patch, it remains highly effective, as shown in Tab.~\ref{tab:trans}. This mainly because the attention can still be misled to focus on the adversarial patch as long as it is perfectly aligned with the input patch. In contrast, if a single pixel is shifted away, the structure of the adversarial perturbation is destroyed due to the misalignment between the input patch of DeiT and the constructed adversarial patch. Additionally, We find that the adversarial patch perturbation can hardly transfer across images or models regardless of the alignment. Details can be found in Appendix I.

\section{Improving ViT Robustness to Adversarial Patch}

Given an input image $\boldsymbol{x}\in \mathbb{R}^{H\times W\times C}$, ViT \cite{dosovitskiy2020image} first reshapes the input $\boldsymbol{x}$ into a sequence of image patches $\{\boldsymbol{x}_i\in \mathbb{R}^{(\frac{H}{P}\cdot \frac{W}{P})\times (P^2\cdot C)}\}_{i=1}^N$ where $P$ is the patch size and $N$ is the number of patches. A class-token patch $\boldsymbol{x}_0$ is concatenated to the patch sequence. A set of self-attention blocks is applied to obtain patch embeddings of the $l$-th block $\{\boldsymbol{x}^l_i\}_{i=0}^N$. The class-token patch embedding of the last block is mapped to the output.

The patch embedding of the $i$-th patch in the $l$-th layer is the weighted sum of all patch embedding $\{\boldsymbol{x}^{l-1}_j\}_{i=0}^N$ of the previous layer. The weights are the attention weights obtained from the attention module. Formally, the patch embedding $\boldsymbol{x}^{l}_i$ is computed with following equation \vspace{-0.3cm}
\begin{equation}
\small
\boldsymbol{x}^{l}_i = \sum_{j=0}^N \alpha_{ij} \cdot \boldsymbol{x}^{l-1}_j, \;\;\;\;\;\;\;\;
\alpha_{ij} = \frac{\exp(Z_{ij})}{\sum_{j=0}^N \exp(Z_{ij})} 
\end{equation}
where $\alpha_{ij}$ is the attention weight that stands for the attention of the $i$-th patch of the $l$-th layer to the $j$-th patch of the $(l$-$1)$-th layer. $Z_{ij}$ is the scaled dot-product between the key of the $j$-th patch and the query of of the $i$-th patch in the $(l$-$1)$-th layer, i.e., the logits before \textit{softmax} attention.

Given a classification task, we denote the patch embedding of the clean image as $\boldsymbol{x}^{*l}_i$. When the $k$-th patch is attacked, the patch embedding of the $i$-th patch in the $l$-th layer deviates from $\boldsymbol{x}^{*l}_i$. The deviation distance is described as 
\begin{equation}
\small
d(\boldsymbol{x}^{l}_i, \boldsymbol{x}^{*l}_i)= \sum_{j=0}^N \alpha_{ij} \cdot \boldsymbol{x}^{l-1}_j - \sum_{j=0}^N \alpha^*_{ij} \cdot \boldsymbol{x}^{l-1}_j,
\end{equation}
where $\alpha^*_{ij}$ is the attention weight corresponding to the clean image. Our analysis shows that the attention is misled to focus on the attacked patch. In other words, $\alpha_{ik}$ is close to $1$, and other attention weights are close to zero.

To address this, we replace the original attention with smoothed attention using temperature scaling in the \textit{softmax} operation. Formally, the smoothed attention is defined as
\vspace{-2mm}
\begin{equation}
\small
\alpha^{\Diamond}_{ij} = \frac{\exp(Z_{ij}/T)}{\sum_{j=0}^N \exp(Z_{ij}/T)},
\end{equation}
where $T (>1)$ is the hyper-parameter that determines the smoothness of the proposed attention. With the smoothed attention, the deviation of the patch embedding from the clean patch embedding is smaller. \vspace{-0.2cm}
\begin{equation}
\small
d(\boldsymbol{x}^{\Diamond l}_i, \boldsymbol{x}^{*l}_i)= \sum_{j=0}^N \alpha^{\Diamond l}_{ij} \cdot \boldsymbol{x}^{l-1}_j - \sum_{j=0}^N \alpha^*_{ij} \cdot \boldsymbol{x}^{l-1}_j < d(\boldsymbol{x}^{l}_i, \boldsymbol{x}^{*l}_i)
\end{equation}

\vspace{-0.2cm}
\begin{wrapfigure}{r}{0.45\textwidth} \vspace{-0.6cm}
    \includegraphics[width=0.46\textwidth]{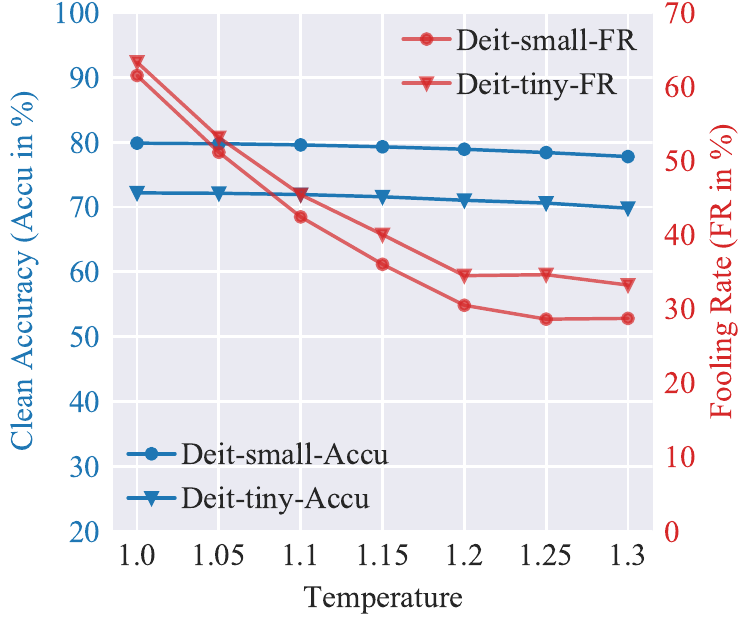} \vspace{-0.6cm}
  \caption{The robustness of ViT can be improved with Smoothed Attention.} \vspace{-0.6cm}
  \label{fig:imp}
\end{wrapfigure} 
We can see that the smoothed attention naturally encourages self-attention not to focus on a single patch. To validate if ViT becomes more robust to adversarial patches, we apply the method to ViT and report the results in Fig.~\ref{fig:imp}. Under different temperatures, the smoothed attention can improve the adversarial robustness of ViT to adversarial patches and rarely reduce the clean accuracy. In addition, the effectiveness of smoothed attention also verifies our understanding of the robustness of ViT in Sec.~\ref{sec:attn_manip}: it is the attention mechanism that causes the vulnerability of ViT against adversarial patch attacks. 

\section{Discussion}
In previous sections, we mainly focus on studying the state-of-the-art patch attack methods on the most primary ViT architecture and ResNet. In this section, we further investigate different variants of model architectures as well as adversarial patch attacks. 

\begin{figure}[t]
   \centering
    \includegraphics[scale=0.2]{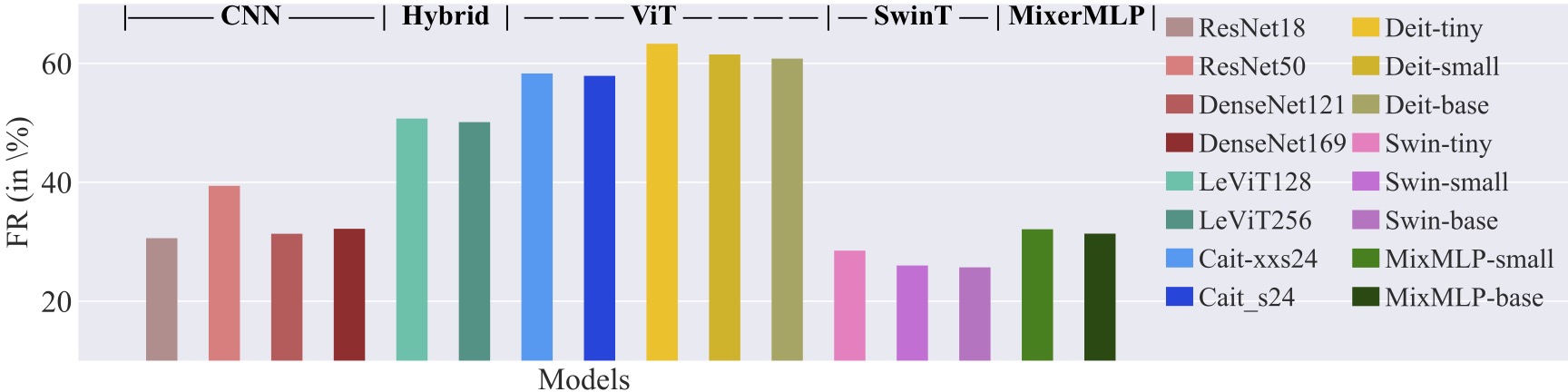}
  \caption{We report Fooling Rates on different versions of ViT, CNN as well as Hybrid architectures under adversarial patch attacks.}
  \label{fig:model_vars}
\end{figure} 

\paragraph{\textrm{\textbf{Different Model Architectures}}} In addition to DeiT and ResNet, we also investigate the robustness of different versions of ViT~\cite{dosovitskiy2020image,touvron2021training,liu2021swin}, CNN~\cite{he2016deep,huang2017densely} as well as Hybrid architectures~\cite{graham2021levit} under adversarial patch attacks. Following the experimental setting in section~\ref{sec:exp_set}, we train all the models and report fooling rate on each model in Fig.~\ref{fig:model_vars}. Four main conclusions can be drawn from the figure.
\begin{enumerate}
    \item  CNN variants are consistently more robust than ViT models.
    \item The robustness of LeViT model~\cite{graham2021levit} with hybrid architecture (\textit{i.e.}, Conv Layers + Self-Attention Blocks) lives somewhere between ViT and CNNs.
    \item Swin Transformers~\cite{liu2021swin} are as robust as CNNs. We conjecture this is because attention cannot be manipulated by a single patch due to hierarchical attention and the shifted windows therein. Specifically, the self-attention in Swin Transformers is conducted on patches within a local region rather than the whole image. In addition, a single patch will interact with patches from different groups in different layers with shifted windows. This makes effective adversarial patches challenging. 
    \item Mixer-MLP~\cite{tolstikhin2021mlp} uses the same patch-based architecture as ViTs and has no attention module. Mixer-base with FR $(31.36)$ is comparable to ResNet and more robust than ViTs. The results further confirm that the vulnerability of ViT can be attributed to self-attention mechanism.
\end{enumerate}

Our proposed attention smoothing by temperature scaling can effectively improve the robustness of DeiT and Levit. However, the improvement on Swin Transformers is tiny due to its architecture design.

\paragraph{\textbf{Different Patch Attacks}} Other than adversarial patch attacks studied previously, we also investigate the robustness of ViT and ResNet against the following variants of adversarial patch attacks.

\paragraph{Imperceptible patch attack} ~In previous sections, we use unbounded local patch attacks where the pixel intensity can be set to any value in the image range $[0, 1]$. The adversarial patches are often visible, as shown in Fig.~\ref{fig:teaser}. In this section, we compare DeiT and ResNet under a popular setting where the adversarial perturbation is imperceptible to humans, bounded by 8/225. In the case of a single patch attack, the attacker achieves FR of 2.9\% on ResNet18 and 11.2\% on DeiT-tiny (see Appendix J for more results). That is: DeiT is still more vulnerable than ResNet when attacked with imperceptible patch perturbation.

\paragraph{Targeted patch attack} ~We also compare DeiT and ResNet under targeted patch attacks, which can be achieved by maximizing the probability of the target class. Specifically, we randomly select a target class other than the ground-truth class for each image. Under a single targeted patch attack, the FR is 15.4\% for ResNet18 vs. 32.3\% for DeiT-tiny, 7.4\% for ResNet50 vs. 24.9\% for DeiT-small. The same conclusion holds: DeiT is more vulnerable than ResNet. Visualization of adversarial patches is in Appendix K.

\paragraph{Patch attack generated with different iterations} ~Following \cite{karmon2018lavan}, we generate adversarial patch attacks with 10k iterations. In this section, we further study the minimum  iterations required to successfully attack the classifier, which is averaged over all patch positions of the misclassified images. We find that the minimum attack iterations on DeiT-tiny is much smaller than that on ResNet18 (65 vs. 342). Similar results on DeiT-small and ResNet50 (294 vs. 455). This further validates DeiT is more vulnerable than ResNet.

\paragraph{ViT-agnostic patch attack} ~In this section, we study ViT-agnostic patch attack where the adversarial patch of the same size as an input patch is placed to a random area of the image. The covered area can involve pixels from multiple input patches. We find that DeiT becomes less vulnerable to adversarial patch attack, \textit{e.g.}, the FR on DeiT-small decreases from 61.5$\%$ to 47.9$\%$. When the adversarial patch is not aligned with the input patch, \textit{i.e.}, only part of patch pixels can be manipulated, the attention of DeiT is less likely to be misled. Under such ViT-agnostic patch attack, ViT is still more vulnerable than ResNet.

\section{Conclusion}
This work starts with an interesting observation on the robustness of ViT to patch perturbations. Namely, vision transformer (e.g., DeiT) is more robust to natural patch corruption than ResNet, whereas it is significantly more vulnerable against adversarial patches. Further, we discover the self-attention mechanism of ViT can effectively ignore natural corrupted patches but be easily misled to adversarial patches to make mistakes. Based on our analysis, we propose attention smoothing to improve the robustness of ViT to adversarial patches, which further validates our developed understanding. We believe this study can help the community better understand the robustness of ViT to patch perturbations.

\bibliographystyle{splncs04}
\bibliography{egbib}

\begin{thebibliography}{10}
\providecommand{\url}[1]{\texttt{#1}}
\providecommand{\urlprefix}{URL }
\providecommand{\doi}[1]{https://doi.org/#1}

\bibitem{abnar2020quantifying}
Abnar, S., Zuidema, W.: Quantifying attention flow in transformers. In: Annual
  Meeting of the Association for Computational Linguistics (ACL) (2020)

\bibitem{aldahdooh2021reveal}
Aldahdooh, A., Hamidouche, W., Deforges, O.: Reveal of vision transformers
  robustness against adversarial attacks. arXiv:2106.03734  (2021)

\bibitem{bai2021transformers}
Bai, Y., Mei, J., Yuille, A., Xie, C.: Are transformers more robust than cnns?
  arXiv:2111.05464  (2021)

\bibitem{benz2021adversarial}
Benz, P., Ham, S., Zhang, C., Karjauv, A., Kweon, I.S.: Adversarial robustness
  comparison of vision transformer and mlp-mixer to cnns. arXiv preprint
  arXiv:2110.02797  (2021)

\bibitem{bhojanapalli2021understanding}
Bhojanapalli, S., Chakrabarti, A., Glasner, D., Li, D., Unterthiner, T., Veit,
  A.: Understanding robustness of transformers for image classification.
  arXiv:2103.14586  (2021)

\bibitem{brown2017adversarial}
Brown, T.B., Man{\'e}, D., Roy, A., Abadi, M., Gilmer, J.: Adversarial patch.
  arXiv:1712.09665v1  (2017)

\bibitem{chen2021crossvit}
Chen, C.F., Fan, Q., Panda, R.: Crossvit: Cross-attention multi-scale vision
  transformer for image classification. arXiv:2103.14899  (2021)

\bibitem{chen2021visformer}
Chen, Z., Xie, L., Niu, J., Liu, X., Wei, L., Tian, Q.: Visformer: The
  vision-friendly transformer. arXiv:2104.12533  (2021)

\bibitem{deng2009imagenet}
Deng, J., Dong, W., Socher, R., Li, L.J., Li, K., Fei-Fei, L.: Imagenet: A
  large-scale hierarchical image database. In: IEEE Conference on Computer
  Vision and Pattern Recognition (CVPR) (2009)

\bibitem{dosovitskiy2020image}
Dosovitskiy, A., Beyer, L., Kolesnikov, A., Weissenborn, D., Zhai, X.,
  Unterthiner, T., Dehghani, M., Minderer, M., Heigold, G., Gelly, S., et~al.:
  An image is worth 16x16 words: Transformers for image recognition at scale.
  arXiv:2010.11929  (2020)

\bibitem{BMVC2016_137}
Fawzi, A., Frossard, P.: Measuring the effect of nuisance variables on
  classifiers. In: Proceedings of the British Machine Vision Conference (BMVC)
  (2016)

\bibitem{fu2021patch}
Fu, Y., Zhang, S., Wu, S., Wan, C., Lin, Y.: Patch-fool: Are vision
  transformers always robust against adversarial perturbations? In:
  International Conference on Learning Representations (2021)

\bibitem{goodfellow2014explaining}
Goodfellow, I.J., Shlens, J., Szegedy, C.: Explaining and harnessing
  adversarial examples. arXiv:1412.6572  (2014)

\bibitem{graham2021levit}
Graham, B., El-Nouby, A., Touvron, H., Stock, P., Joulin, A., J{\'e}gou, H.,
  Douze, M.: Levit: a vision transformer in convnet's clothing for faster
  inference. arXiv:2104.01136  (2021)

\bibitem{han2021transformer}
Han, K., Xiao, A., Wu, E., Guo, J., Xu, C., Wang, Y.: Transformer in
  transformer. arXiv:2103.00112  (2021)

\bibitem{he2016deep}
He, K., Zhang, X., Ren, S., Sun, J.: Deep residual learning for image
  recognition. In: IEEE Conference on Computer Vision and Pattern Recognition
  (CVPR) (2016)

\bibitem{hendrycks2019benchmarking}
Hendrycks, D., Dietterich, T.: Benchmarking neural network robustness to common
  corruptions and perturbations. In: International Conference on Learning
  Representations (ICLR) (2019)

\bibitem{hu2021inheritance}
Hu, H., Lu, X., Zhang, X., Zhang, T., Sun, G.: Inheritance attention
  matrix-based universal adversarial perturbations on vision transformers. IEEE
  Signal Processing Letters  \textbf{28},  1923--1927 (2021)

\bibitem{huang2017densely}
Huang, G., Liu, Z., Van Der~Maaten, L., Weinberger, K.Q.: Densely connected
  convolutional networks. In: Proceedings of the IEEE conference on computer
  vision and pattern recognition. pp. 4700--4708 (2017)

\bibitem{joshi2021adversarial}
Joshi, A., Jagatap, G., Hegde, C.: Adversarial token attacks on vision
  transformers. arXiv:2110.04337  (2021)

\bibitem{karmon2018lavan}
Karmon, D., Zoran, D., Goldberg, Y.: Lavan: Localized and visible adversarial
  noise. In: International Conference on Machine Learning (ICML) (2018)

\bibitem{kolesnikov2020big}
Kolesnikov, A., Beyer, L., Zhai, X., Puigcerver, J., Yung, J., Gelly, S.,
  Houlsby, N.: Big transfer (bit): General visual representation learning. In:
  European Conference on Computer Vision (ECCV) (2020)

\bibitem{liu2019perceptual}
Liu, A., Liu, X., Fan, J., Ma, Y., Zhang, A., Xie, H., Tao, D.:
  Perceptual-sensitive gan for generating adversarial patches. In: AAAI (2019)

\bibitem{liu2020bias}
Liu, A., Wang, J., Liu, X., Cao, B., Zhang, C., Yu, H.: Bias-based universal
  adversarial patch attack for automatic check-out. In: European conference on
  computer vision. pp. 395--410. Springer (2020)

\bibitem{liu2021swin}
Liu, Z., Lin, Y., Cao, Y., Hu, H., Wei, Y., Zhang, Z., Lin, S., Guo, B.: Swin
  transformer: Hierarchical vision transformer using shifted windows.
  arXiv:2103.14030  (2021)

\bibitem{luo2021generating}
Luo, J., Bai, T., Zhao, J.: Generating adversarial yet inconspicuous patches
  with a single image (student abstract). In: Proceedings of the AAAI
  Conference on Artificial Intelligence. vol.~35, pp. 15837--15838 (2021)

\bibitem{madry2017towards}
Madry, A., Makelov, A., Schmidt, L., Tsipras, D., Vladu, A.: Towards deep
  learning models resistant to adversarial attacks. In: arXiv:1706.06083 (2017)

\bibitem{mahmood2021robustness}
Mahmood, K., Mahmood, R., Van~Dijk, M.: On the robustness of vision
  transformers to adversarial examples. In: Proceedings of the IEEE/CVF
  International Conference on Computer Vision. pp. 7838--7847 (2021)

\bibitem{mao2021towards}
Mao, X., Qi, G., Chen, Y., Li, X., Duan, R., Ye, S., He, Y., Xue, H.: Towards
  robust vision transformer. arXiv:2105.07926  (2021)

\bibitem{mao2021rethinking}
Mao, X., Qi, G., Chen, Y., Li, X., Ye, S., He, Y., Xue, H.: Rethinking the
  design principles of robust vision transformer. arXiv:2105.07926  (2021)

\bibitem{metzen2021meta}
Metzen, J.H., Finnie, N., Hutmacher, R.: Meta adversarial training against
  universal patches. arXiv preprint arXiv:2101.11453  (2021)

\bibitem{mu2021defending}
Mu, N., Wagner, D.: Defending against adversarial patches with robust
  self-attention. In: ICML 2021 Workshop on Uncertainty and Robustness in Deep
  Learning (2021)

\bibitem{naseer2021intriguing}
Naseer, M., Ranasinghe, K., Khan, S., Hayat, M., Khan, F.S., Yang, M.H.:
  Intriguing properties of vision transformers. arXiv:2105.10497  (2021)

\bibitem{naseer2021improving}
Naseer, M., Ranasinghe, K., Khan, S., Khan, F.S., Porikli, F.: On improving
  adversarial transferability of vision transformers. arXiv:2106.04169  (2021)

\bibitem{papernot2016limitations}
Papernot, N., McDaniel, P., Jha, S., Fredrikson, M., Celik, Z.B., Swami, A.:
  The limitations of deep learning in adversarial settings. In: 2016 IEEE
  European symposium on security and privacy (EuroS\&P) (2016)

\bibitem{paul2021vision}
Paul, S., Chen, P.Y.: Vision transformers are robust learners. arXiv:2105.07581
   (2021)

\bibitem{qian2020visually}
Qian, Y., Wang, J., Wang, B., Zeng, S., Gu, Z., Ji, S., Swaileh, W.: Visually
  imperceptible adversarial patch attacks on digital images. arXiv preprint
  arXiv:2012.00909  (2020)

\bibitem{qin2021understanding}
Qin, Y., Zhang, C., Chen, T., Lakshminarayanan, B., Beutel, A., Wang, X.:
  Understanding and improving robustness of vision transformers through
  patch-based negative augmentation. arXiv preprint arXiv:2110.07858  (2021)

\bibitem{salman2021certified}
Salman, H., Jain, S., Wong, E., Madry, A.: Certified patch robustness via
  smoothed vision transformers. arXiv:2110.07719  (2021)

\bibitem{selvaraju2017grad}
Selvaraju, R.R., Cogswell, M., Das, A., Vedantam, R., Parikh, D., Batra, D.:
  Grad-cam: Visual explanations from deep networks via gradient-based
  localization. In: ICCV (2017)

\bibitem{shao2021adversarial}
Shao, R., Shi, Z., Yi, J., Chen, P.Y., Hsieh, C.J.: On the adversarial
  robustness of visual transformers. arXiv:2103.15670  (2021)

\bibitem{shi2021decision}
Shi, Y., Han, Y.: Decision-based black-box attack against vision transformers
  via patch-wise adversarial removal. arXiv preprint arXiv:2112.03492  (2021)

\bibitem{shrikumar2017learning}
Shrikumar, A., Greenside, P., Kundaje, A.: Learning important features through
  propagating activation differences. In: International Conference on Machine
  Learning (ICML) (2017)

\bibitem{szegedy2013intriguing}
Szegedy, C., Zaremba, W., Sutskever, I., Bruna, J., Erhan, D., Goodfellow, I.,
  Fergus, R.: Intriguing properties of neural networks. International
  Conference on Learning Representations (ICLR)  (2014)

\bibitem{tang2021robustart}
Tang, S., Gong, R., Wang, Y., Liu, A., Wang, J., Chen, X., Yu, F., Liu, X.,
  Song, D., Yuille, A., et~al.: Robustart: Benchmarking robustness on
  architecture design and training techniques. arXiv preprint arXiv:2109.05211
  (2021)

\bibitem{tolstikhin2021mlp}
Tolstikhin, I., Houlsby, N., Kolesnikov, A., Beyer, L., Zhai, X., Unterthiner,
  T., Yung, J., Keysers, D., Uszkoreit, J., Lucic, M., et~al.: Mlp-mixer: An
  all-mlp architecture for vision. In: arXiv:2105.01601 (2021)

\bibitem{touvron2021training}
Touvron, H., Cord, M., Douze, M., Massa, F., Sablayrolles, A., J{\'e}gou, H.:
  Training data-efficient image transformers \& distillation through attention.
  In: International Conference on Machine Learning (ICML) (2021)

\bibitem{wang2021universal}
Wang, J., Liu, A., Bai, X., Liu, X.: Universal adversarial patch attack for
  automatic checkout using perceptual and attentional bias. IEEE Transactions
  on Image Processing  \textbf{31},  598--611 (2021)

\bibitem{wu2020visual}
Wu, B., Xu, C., Dai, X., Wan, A., Zhang, P., Yan, Z., Tomizuka, M., Gonzalez,
  J., Keutzer, K., Vajda, P.: Visual transformers: Token-based image
  representation and processing for computer vision. arXiv:2006.03677  (2020)

\bibitem{xiao2021early}
Xiao, T., Singh, M., Mintun, E., Darrell, T., Doll{\'a}r, P., Girshick, R.:
  Early convolutions help transformers see better. arXiv:2106.14881  (2021)

\bibitem{yu2021mia}
Yu, Z., Fu, Y., Li, S., Li, C., Lin, Y.: Mia-former: Efficient and robust
  vision transformers via multi-grained input-adaptation. arXiv preprint
  arXiv:2112.11542  (2021)

\bibitem{zeiler2014visualizing}
Zeiler, M.D., Fergus, R.: Visualizing and understanding convolutional networks.
  In: European conference on computer vision (2014)

\bibitem{zhou2016learning}
Zhou, B., Khosla, A., Lapedriza, A., Oliva, A., Torralba, A.: Learning deep
  features for discriminative localization. In: IEEE Conference on Computer
  Vision and Pattern Recognition (CVPR) (2016)

\end{thebibliography}

\newpage
\appendix

\section{Training Setting Affect Model Robustness}
\label{app:fair_model}
We train ResNet18 on CIFAR10 in the standard setting \cite{he2016deep}. To study the impact of training settings on model robustness, we train models with different input sizes (i.e., 32, 48, 64), with or without Weight Standardization and Group Normalization to regularize the training process. The foolong rate of single patch attack is reported. Especially, with our experiments, we find that Weight Standardization and Group Normalization can have a significant impact on model robustness (See Tab.~\ref{tab:unfair_factors}). The two techniques are applied in BiT \cite{kolesnikov2020big} to improve its performance. However, they are not applied to standard ViT and DeiT training settings. Hence, the robustness difference between ViT and BiT cannot be attributed to the difference between model architectures.

Note that a comprehensive study of the relationship between all factors of training and model adversarial robustness is out of the scope of this paper. We aim to point out that these factors can have an impact on model robustness to different extents. The robustness difference cannot be blindly attributed to the difference of model architectures. We need to build new fair base models to study the robustness of ResNet and ViT.

\begin{table}[!h]
\caption{Study of the training factors on the relation to model robustness: While the input size has minor impact on model robustness in the first tabular, Weight Standardization (WS) and Group Normalization (GN) can change model robustness significantly in the second tabular.}
\label{tab:unfair_factors}
\small
\setlength\tabcolsep{0.14cm}
\centering
\begin{tabular}{c ccc}
\toprule
Model  & \multicolumn{3}{c}{Input Size} \\
\midrule
ResNet18  & 32 & 48 & 64 \\
\midrule
Clean Accuracy     & 93.4 & 93.8 & 93.7    \\
FR of Patch Attack  & 35.9 & 42.2 & 39.2  \\
\bottomrule
\end{tabular} \vspace{0.1cm}

\begin{tabular}{c cccc}
\toprule
Model  & \multicolumn{4}{c}{Training Techniques} \\
\midrule
ResNet18  & No & WS & GN & WS + GN \\
\midrule
Clean Accu & 93.4 & 93.6 & 92.0 & 93.8 \\
Patch Attack FR  & 35.9  & 51.3 & 52.6 & 71.1 \\
\bottomrule
\end{tabular}
\end{table}

\begin{table}[t]
\caption{Fair base models. DeiT and counter-part ResNet are trained with the exact same setting. Two models of each pair achieve similar clean accuracy with comparable model sizes.}
\label{tab:fair_models}
\begin{center}
\footnotesize
\setlength\tabcolsep{0.6cm}
\begin{tabular}{ccc}
\toprule
Model  & Model Size & Clean Accuracy \\
\midrule
ResNet50   & 25M  & 78.79  \\
DeiT-small & 22M  & 79.85 \\
\midrule
ResNet18  & 12M  & 69.39   \\
DeiT-tiny & 5M   & 72.18  \\
\bottomrule
\end{tabular}
\end{center}
\end{table}

\section{Natural Patch Corruption with Different Levels and Types}
\label{app:corr_type}
\begin{figure*}[!ht]
    \centering
    \begin{subfigure}[b]{0.66\textwidth}
    \includegraphics[width=\textwidth]{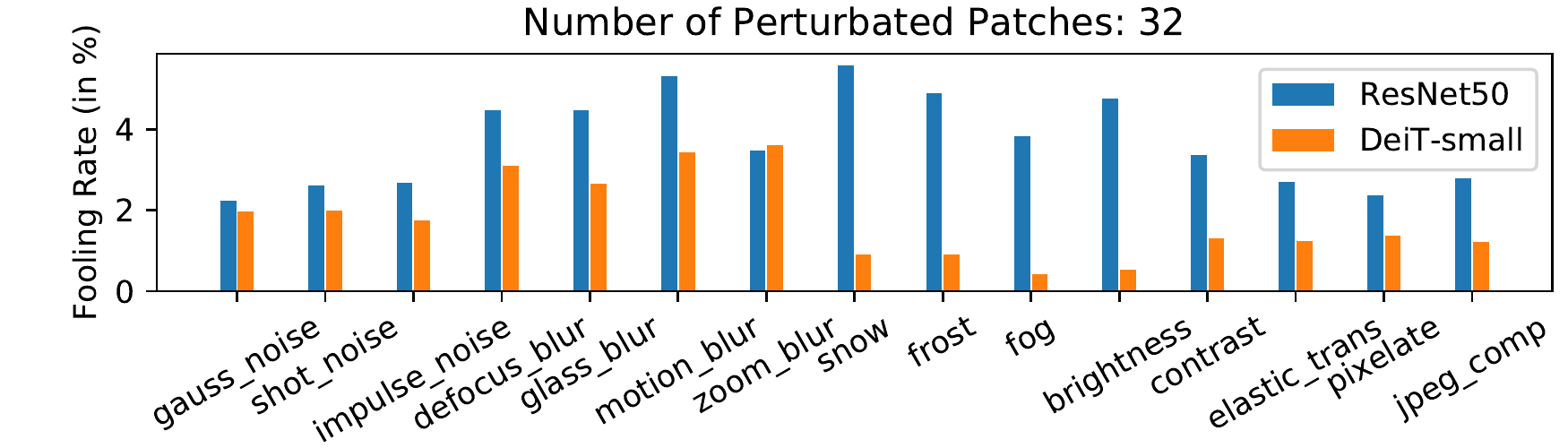}
    \end{subfigure}
    \begin{subfigure}[b]{0.66\textwidth}
        \includegraphics[width=\textwidth]{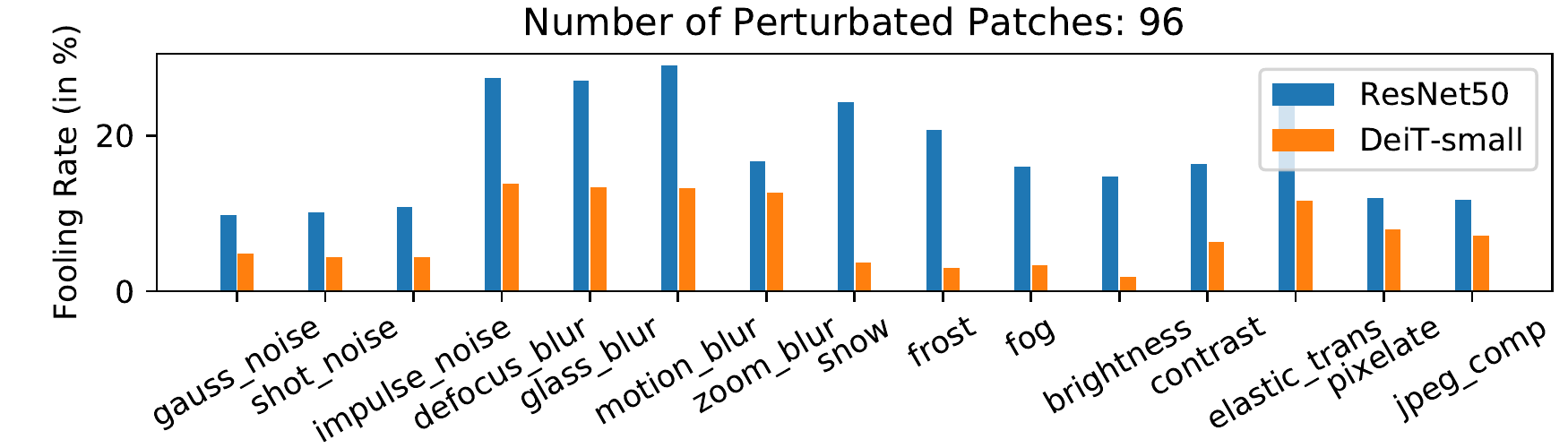}
    \end{subfigure}
    \begin{subfigure}[b]{0.66\textwidth}
        \includegraphics[width=\textwidth]{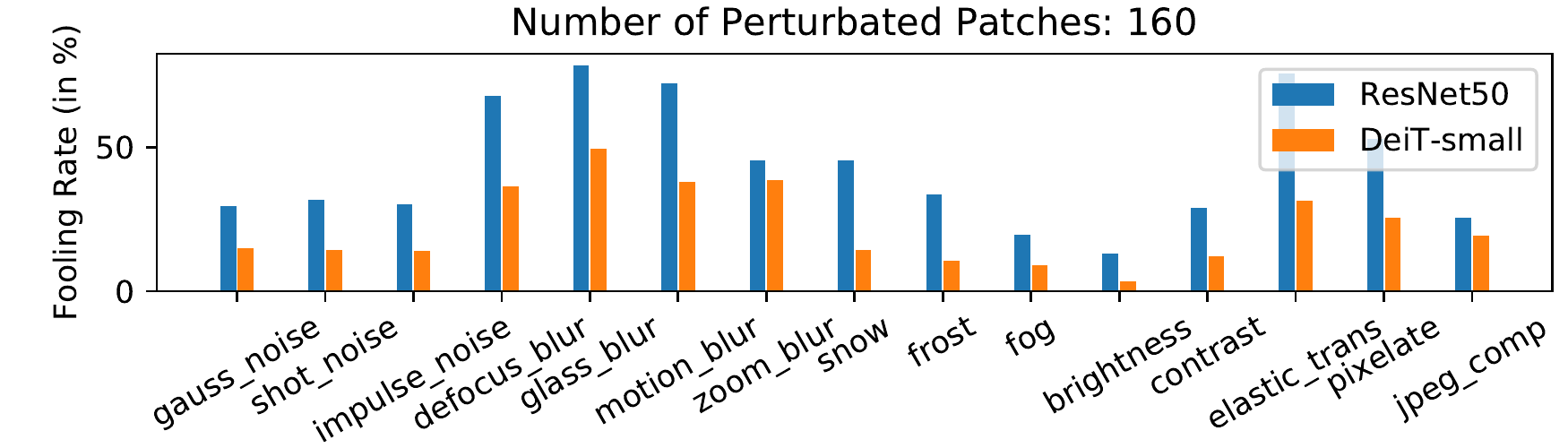}
    \end{subfigure}
    \caption{Comparison of ResNet50 and Deit-small on Naturally Corrupted Patches}
    \label{fig:nat_corrupt_res50_deit_small}
\end{figure*}

\begin{figure*}[!ht]
    \centering
    \begin{subfigure}[b]{0.66\textwidth}
    \includegraphics[width=\textwidth]{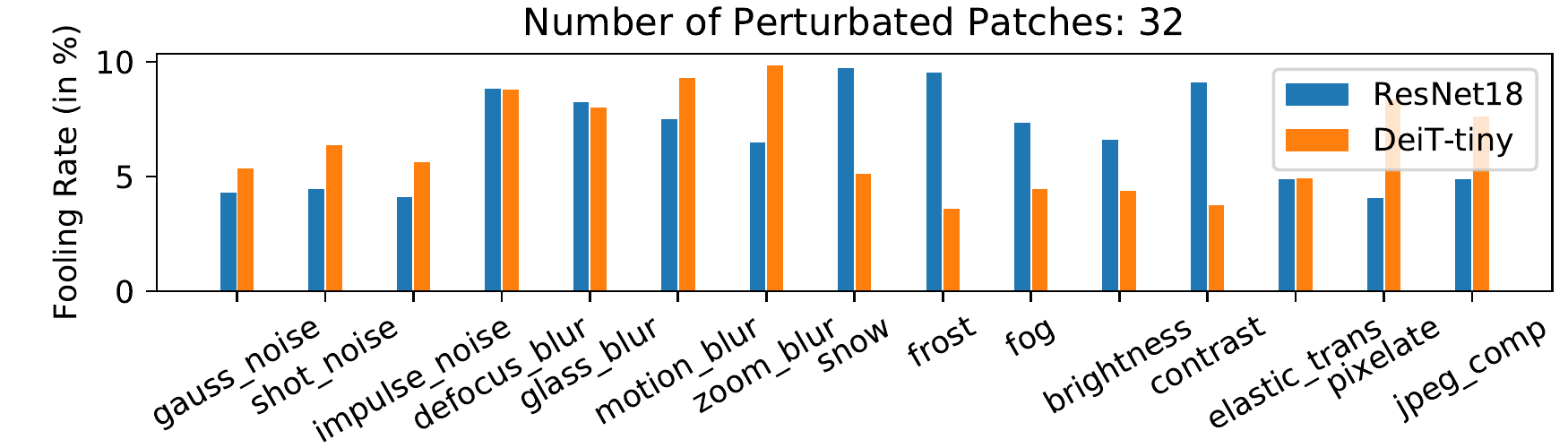}
    \end{subfigure}
    \begin{subfigure}[b]{0.66\textwidth}
        \includegraphics[width=\textwidth]{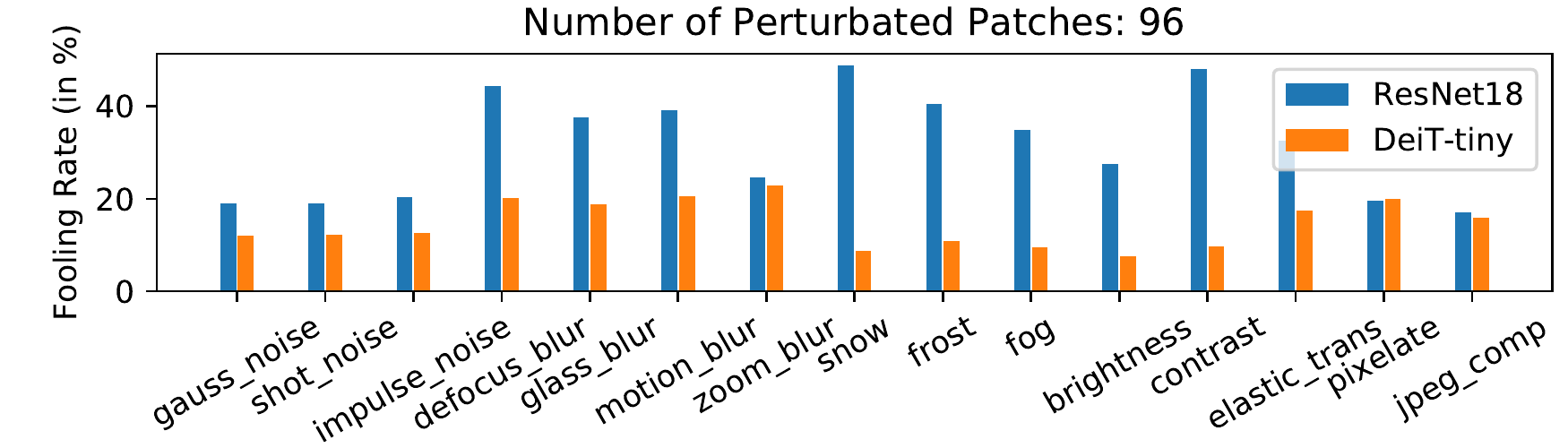}
    \end{subfigure}
    \begin{subfigure}[b]{0.66\textwidth}
        \includegraphics[width=\textwidth]{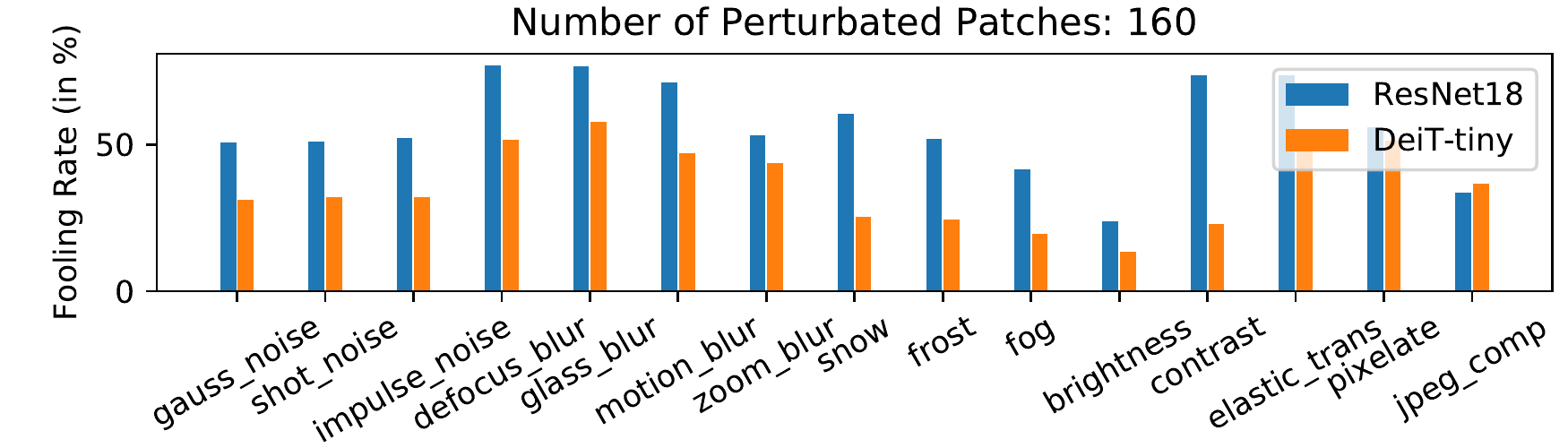}
    \end{subfigure}
    \caption{Comparison of ResNet18 and Deit-tiny on Naturally Corrupted Patches}
    \label{fig:nat_corrupt_res18_deit_tiny}
\end{figure*}

Models can show different robustness when the inputs are corrupted with different natural noise types. To better evaluate the model robustness to natural corruption, the work \cite{hendrycks2019benchmarking} summarizes 15 common natural corruption types. The averaged score is used as an indicator of model robustness. In this appendix section, we show more details of model robustness to different noise types. As show in Fig.~\ref{fig:nat_corrupt_res50_deit_small} and~\ref{fig:nat_corrupt_res18_deit_tiny}, The FR on DeiT is lower than on ResNet. We conclude that DeiT is more robust than ResNet to natural patch corruption.

\begin{figure*}[!ht]
    \centering
    \begin{subfigure}[b]{0.66\textwidth}
    \includegraphics[width=\textwidth]{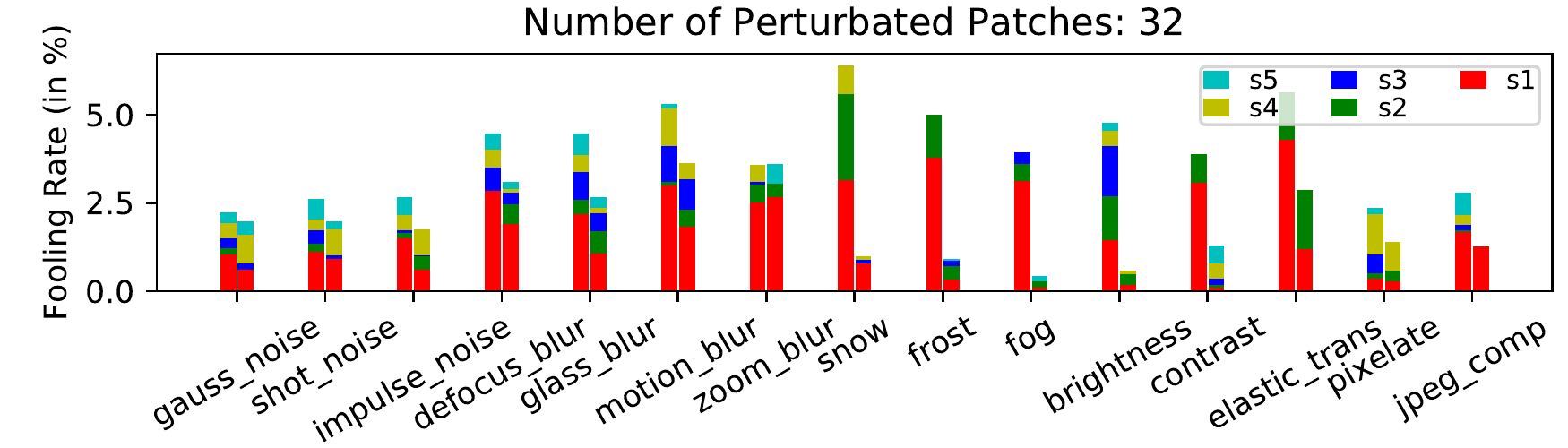}
    \end{subfigure}
    \begin{subfigure}[b]{0.66\textwidth}
        \includegraphics[width=\textwidth]{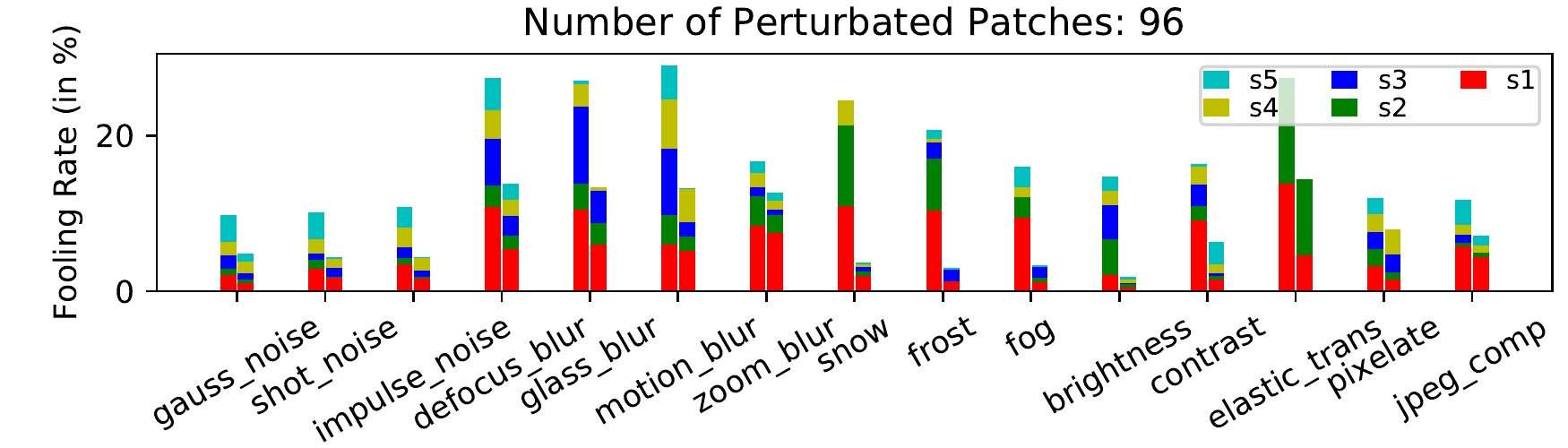}
    \end{subfigure}
    \begin{subfigure}[b]{0.66\textwidth}
        \includegraphics[width=\textwidth]{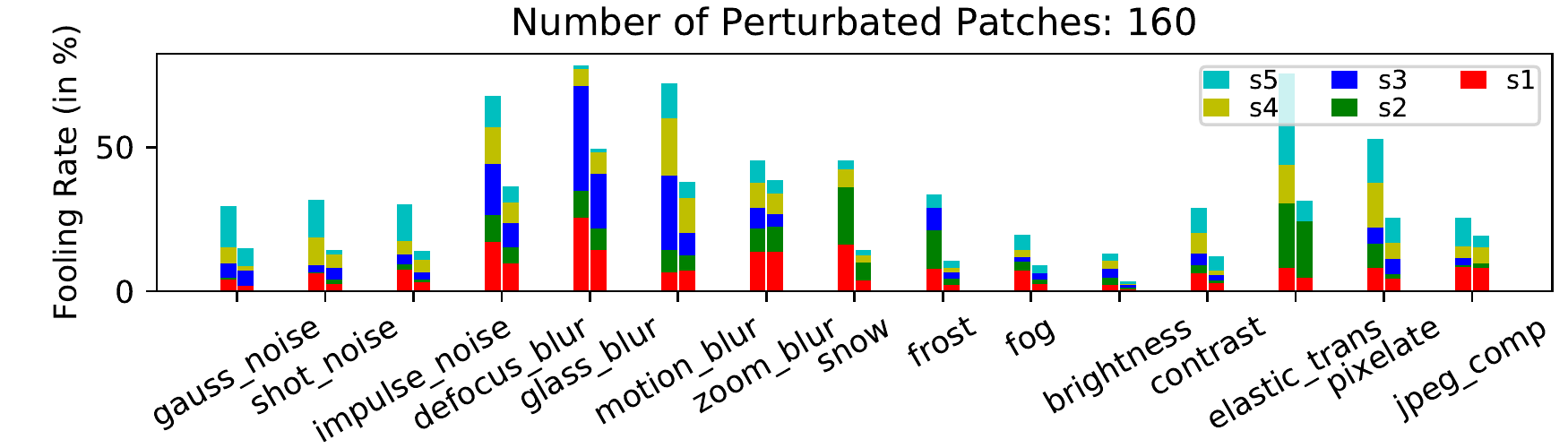}
    \end{subfigure}
    \caption{Comparison of ResNet50 and Deit-small on Patches Corrupted with Different Levels}
    \label{fig:nat_corrupt_res50_deit_small_levels}
\end{figure*}

\begin{figure*}[!ht]
    \centering
    \begin{subfigure}[b]{0.66\textwidth}
    \includegraphics[width=\textwidth]{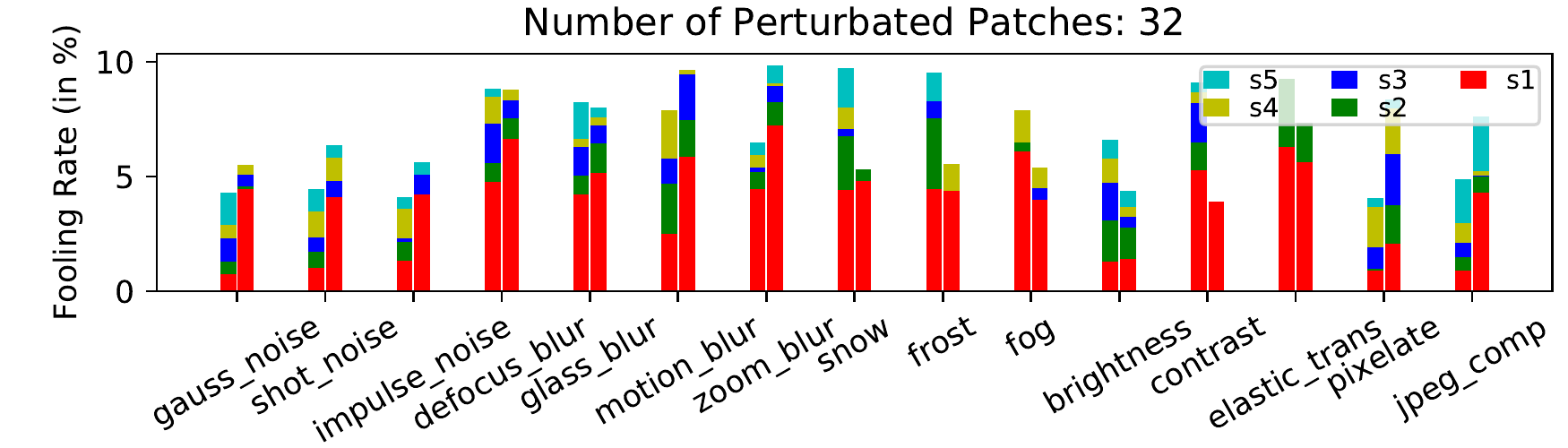}
    \end{subfigure}
    \begin{subfigure}[b]{0.66\textwidth}
        \includegraphics[width=\textwidth]{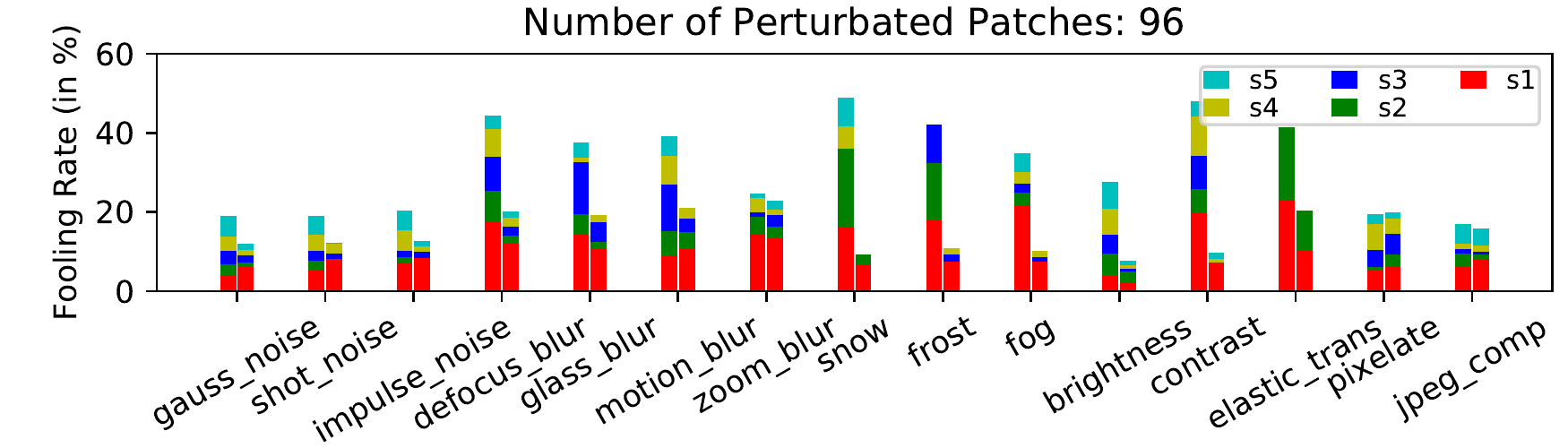}
    \end{subfigure}
    \begin{subfigure}[b]{0.66\textwidth}
        \includegraphics[width=\textwidth]{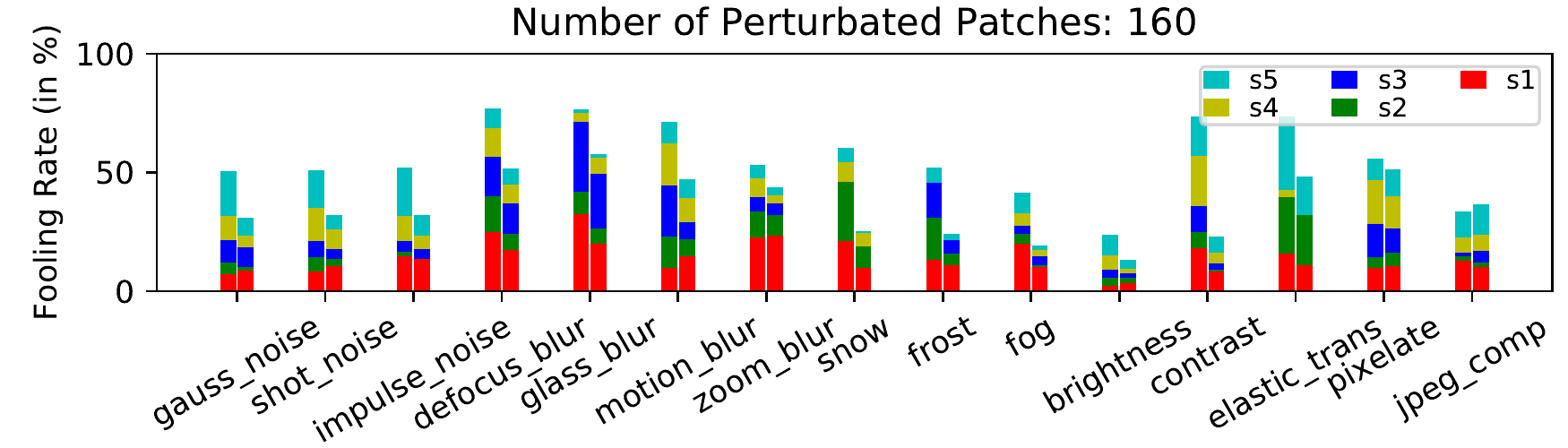}
    \end{subfigure}
    \caption{Comparison of ResNet18 and Deit-tiny on Patches Corrupted with Different Levels}
    \label{fig:nat_corrupt_res18_deit_tiny_levels}
\end{figure*}

Furthermore, we also investigate the model robustness in terms of different noise levels. As shown in Fig.~\ref{fig:nat_corrupt_res50_deit_small_levels} and~\ref{fig:nat_corrupt_res18_deit_tiny_levels}. The different colors stand for different noise level. S1-S5 corresponds to the natural corruption severity from 1 to 5. In each noise type, the left bar corresponds to ResNet variants and the right one to DeiT variants. We can observe that DeiT show lower FR in each severity level. Namely, the conclusion drawn above also holds across different noise levels.

\section{Gradient Visualization of Adversarial Images under Patch Attack}
\label{app:gradient_vis}
We first get the absolute value of gradient received by input and sum them across the channel dimension. The final values are mapped into gray image scale. We also mark the adversarial patch with a blue bounding box in the visualized gradient maps.

\begin{figure*}[!ht]
    \centering
        \includegraphics[width=0.8\textwidth]{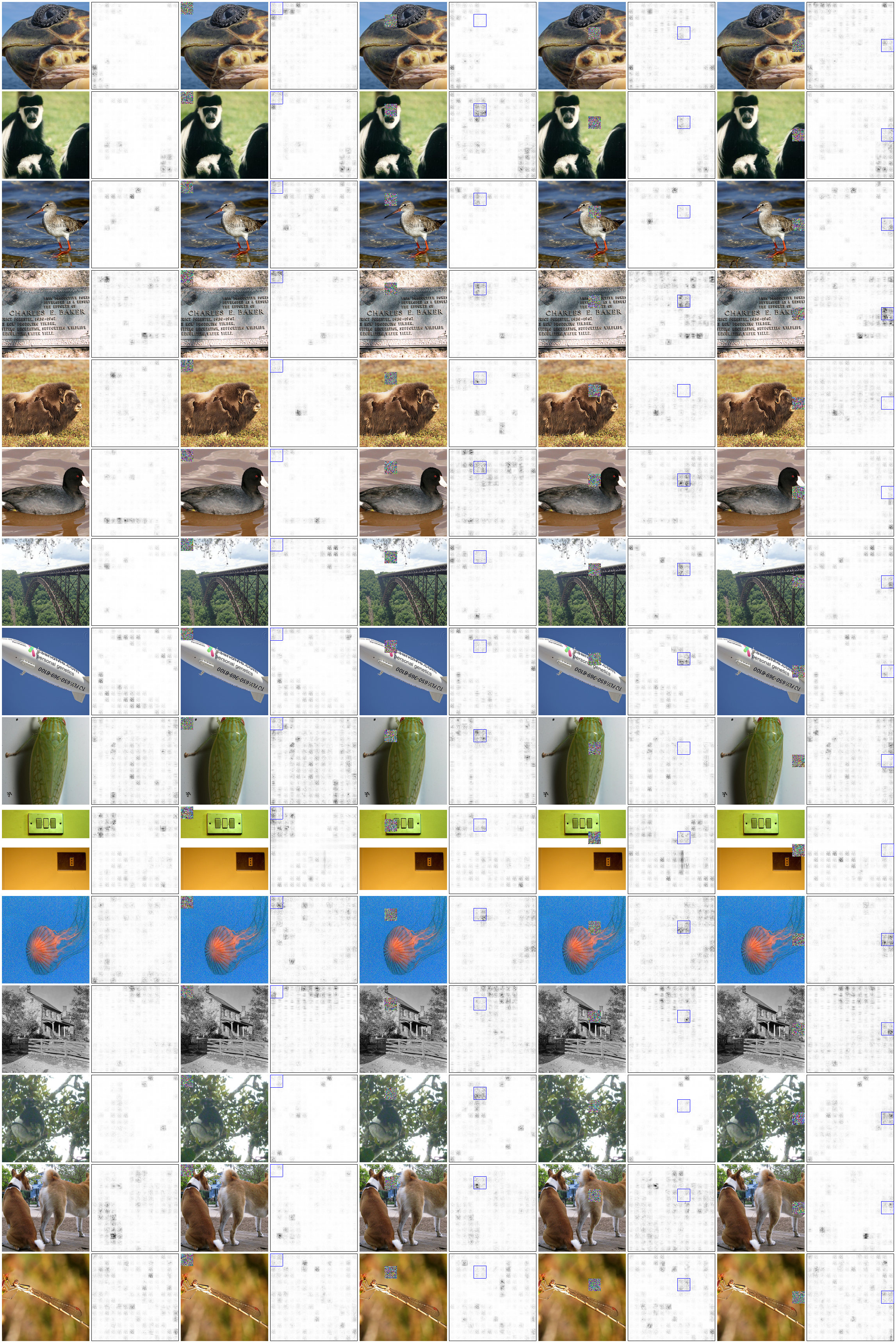}
    \caption{Gradient Visualization on DeiT-small with Attack Patch size of 32}
    \label{fig:grad_deit_p32}
\end{figure*}

\begin{figure*}[!ht]
    \centering
        \includegraphics[width=0.8\textwidth]{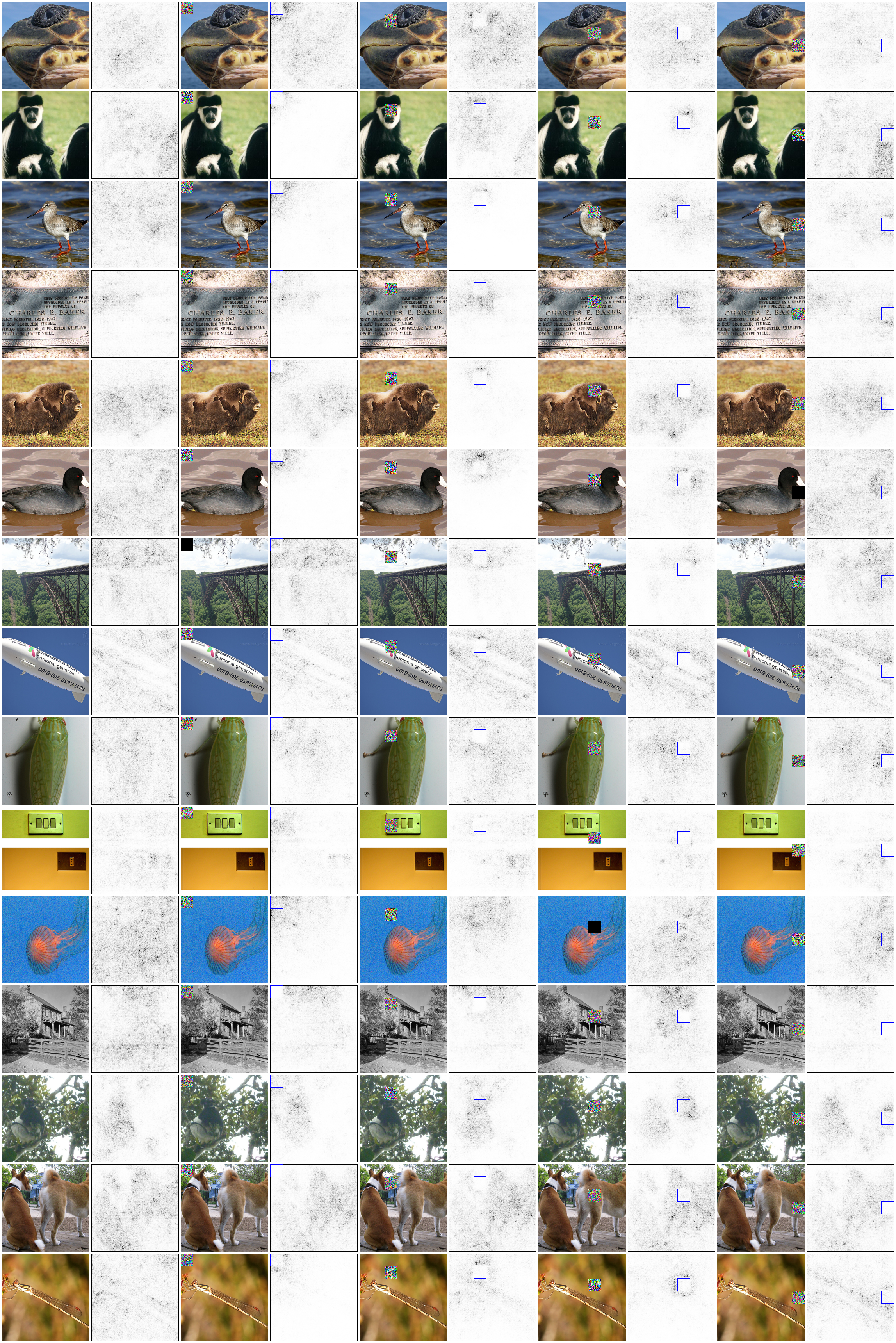}
    \caption{Gradient Visualization on ResNet50 with Attack Patch size of 32}
    \label{fig:grad_res_p32}
\end{figure*}

\begin{figure*}[!ht]
    \centering
        \includegraphics[width=0.8\textwidth]{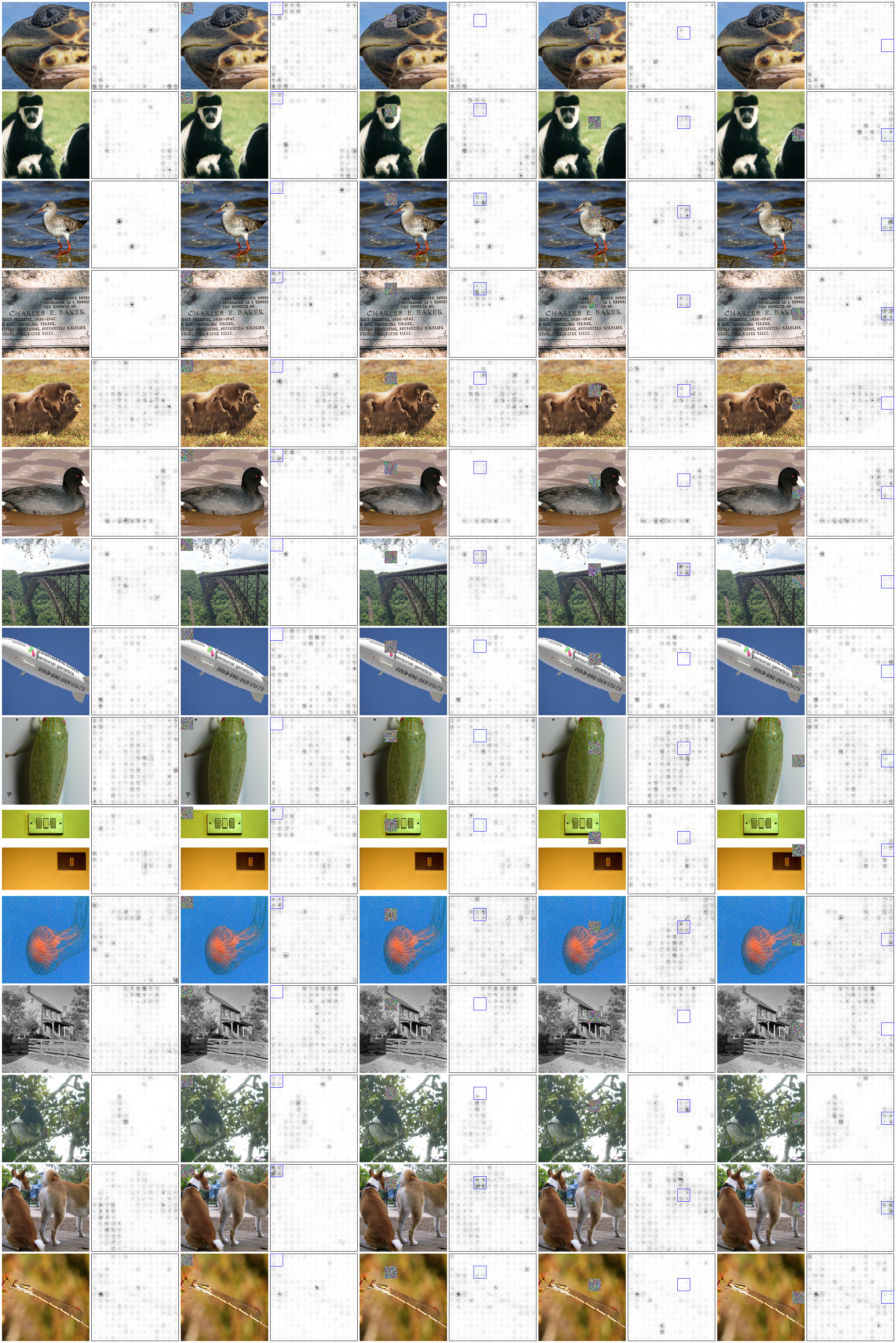}
    \caption{Gradient Visualization on DeiT-tiny with Attack Patch size of 32}
    \label{fig:grad_deit_p16}
\end{figure*}

\begin{figure*}[!ht]
    \centering
        \includegraphics[width=0.8\textwidth]{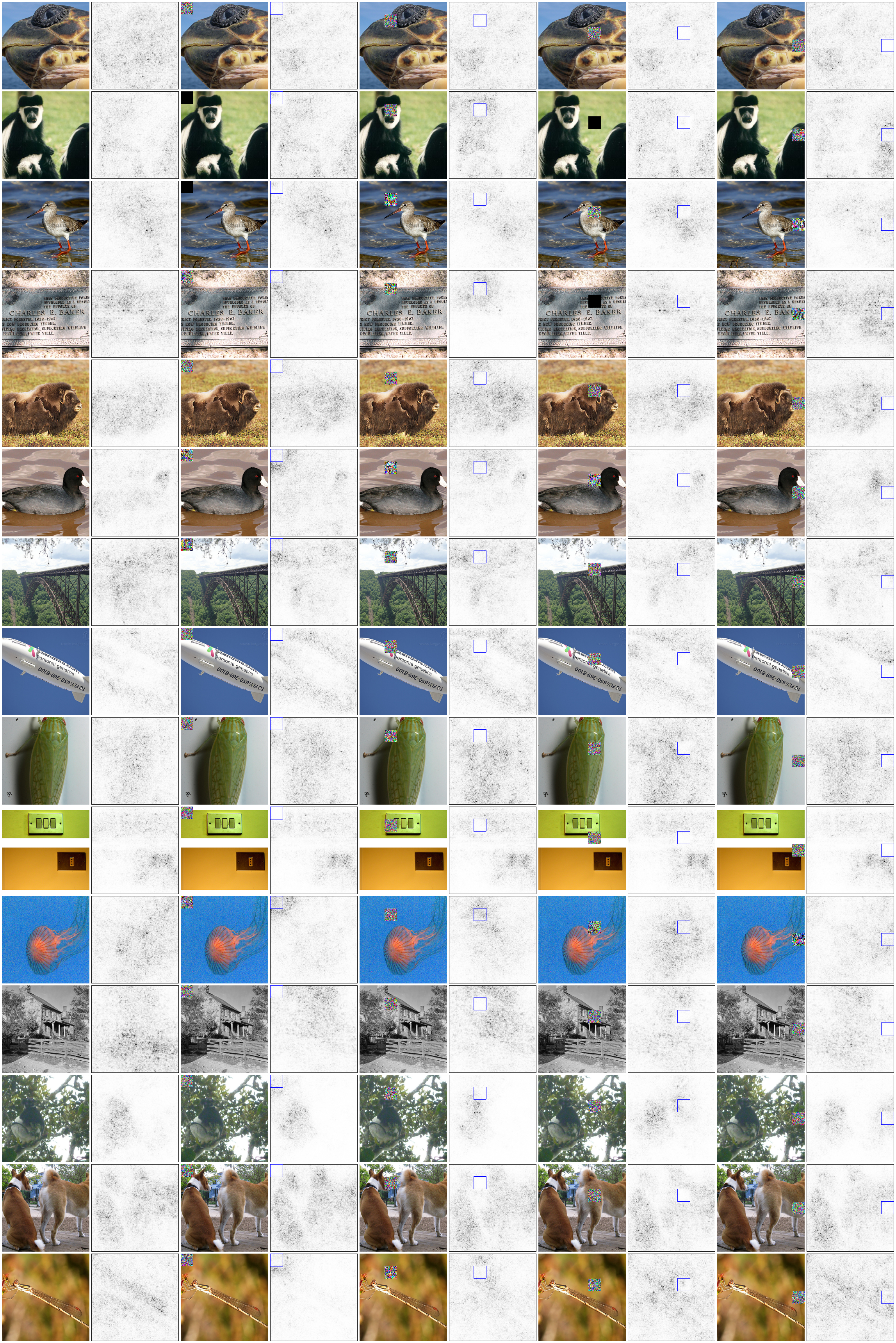}
    \caption{Gradient Visualization on ResNet18 with Attack Patch size of 32}
    \label{fig:grad_res_p16}
\end{figure*}

The adversarial patch noises with different patch sizes (i.e., P=16 and P=32) are shown on DeiT and ResNet in Fig.~\ref{fig:grad_deit_p32},~\ref{fig:grad_res_p32},~\ref{fig:grad_deit_p16}, and~\ref{fig:grad_res_p16}. In each row of these figures, we fist show the clean image and visualize the gradients of inputs as a mask on the image. Then, we show the images with patch noises on different patch positions, and the gradient masks are also shown following the corresponding adversarial images.

\begin{figure*}[t]
    \centering
    \begin{subfigure}[b]{0.4\textwidth}
        \includegraphics[width=\textwidth]{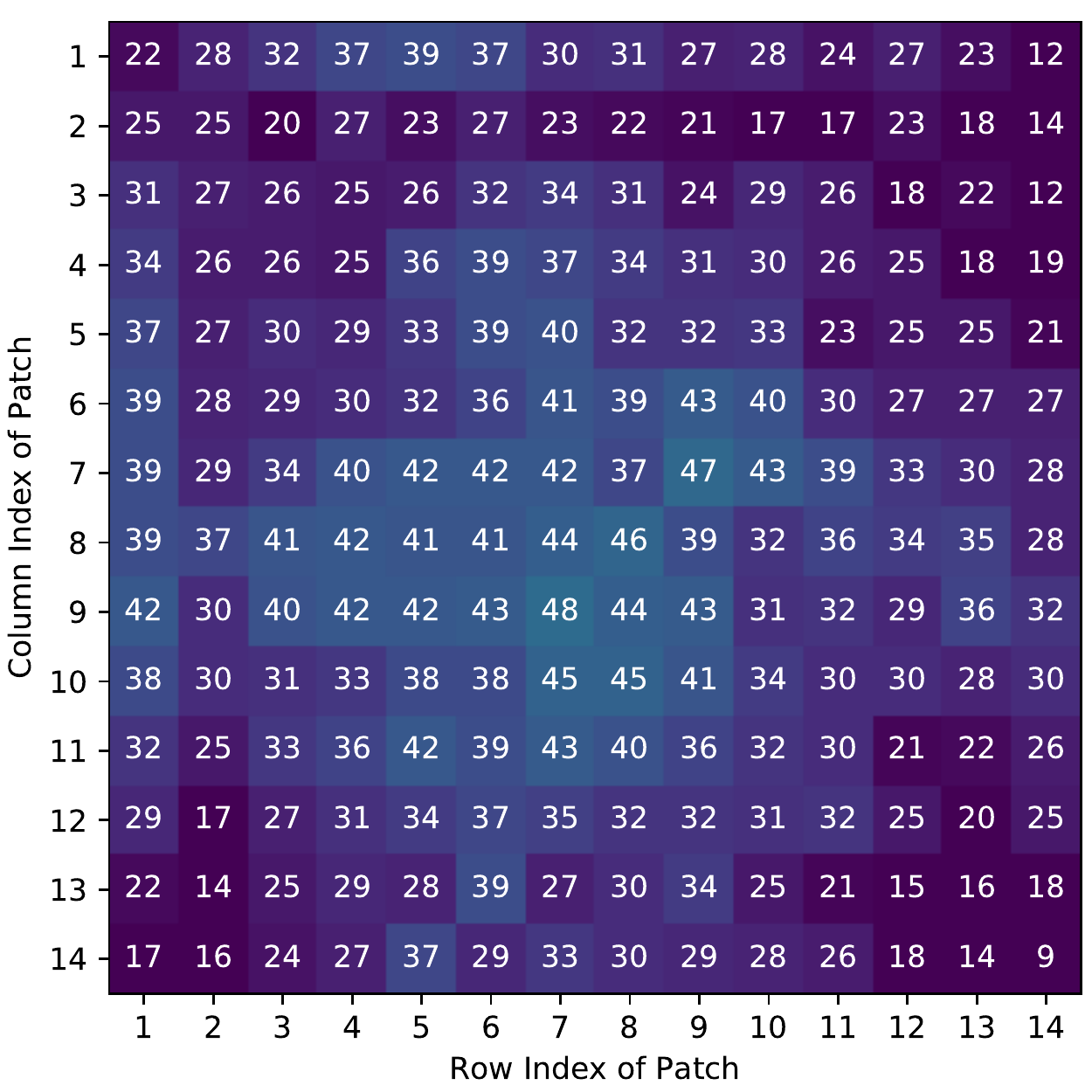}
    \caption{Adversarial Patch Attack FRs on ResNet50}
    \end{subfigure} \hspace{0.2cm}
    \begin{subfigure}[b]{0.4\textwidth}
        \includegraphics[width=\textwidth]{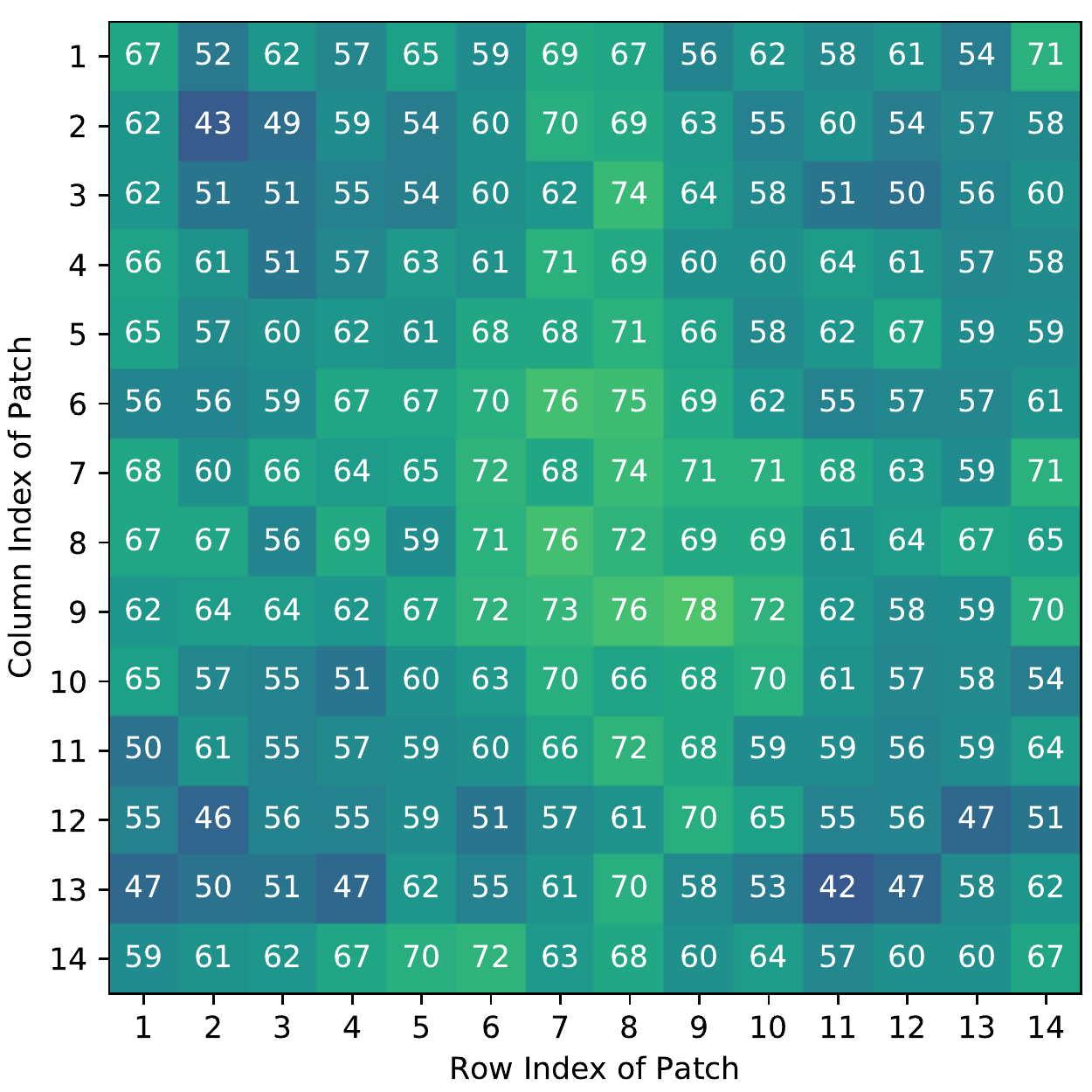}
        \caption{Adversarial Patch Attack FRs on DeiT-small}
    \end{subfigure}
    \caption{Patch Attack FR (in \%) in each patch position is visualized on ResNet50 and DeiT-small.}
    \label{fig:FR_pattern_small_res50}
\end{figure*}

\section{More Figures of Attention on Different Patch Sizes and Positions}
\label{app:rollout_feat}
\begin{figure*}[!ht]
    \centering
        \includegraphics[width=0.8\textwidth]{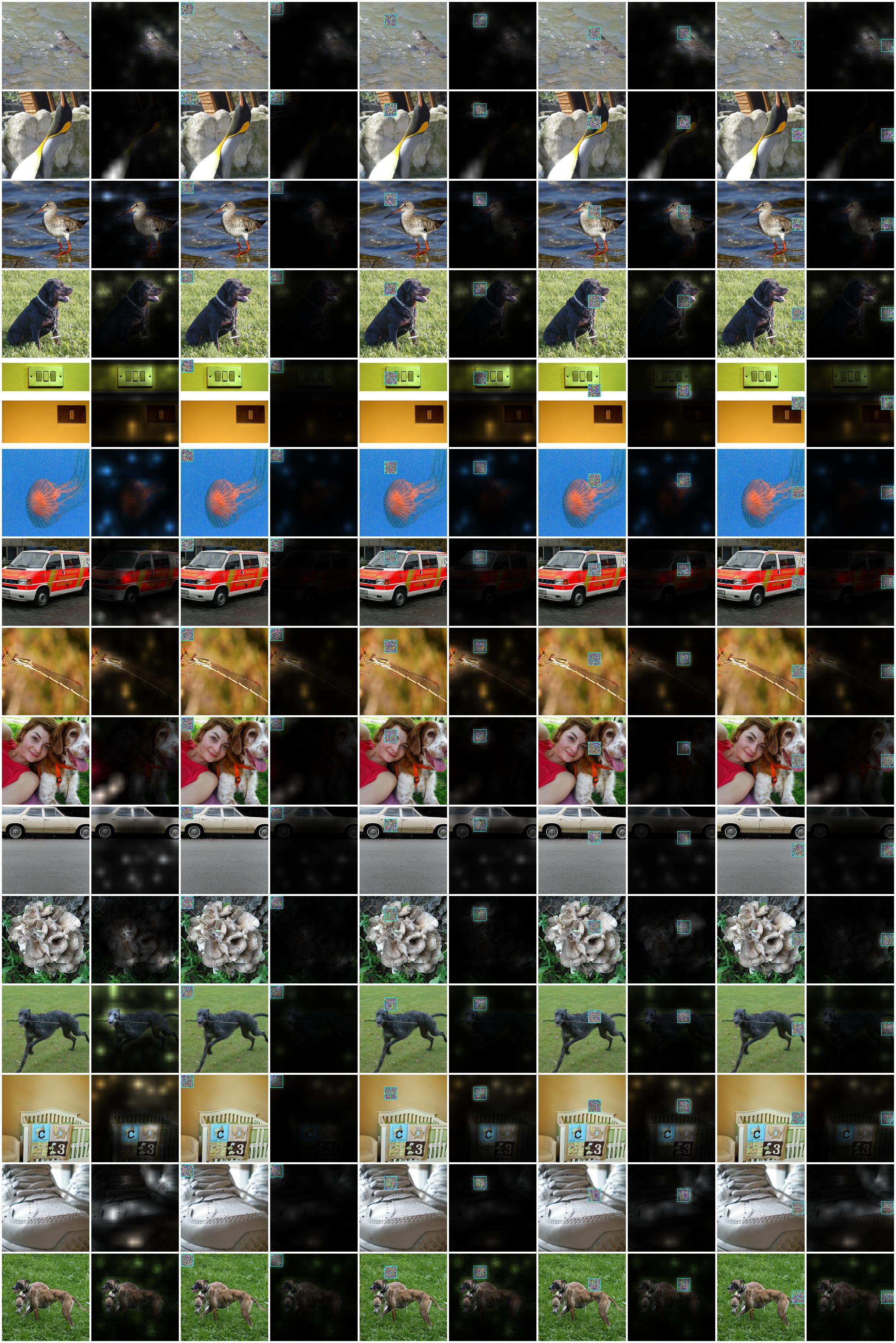}
    \caption{Rollout Attention on DeiT-small with Attack Patch size of 32 on Adversarial Images}
    \label{fig:attn_deit_p32}
\end{figure*}

\begin{figure*}[!ht]
    \centering
        \includegraphics[width=0.8\textwidth]{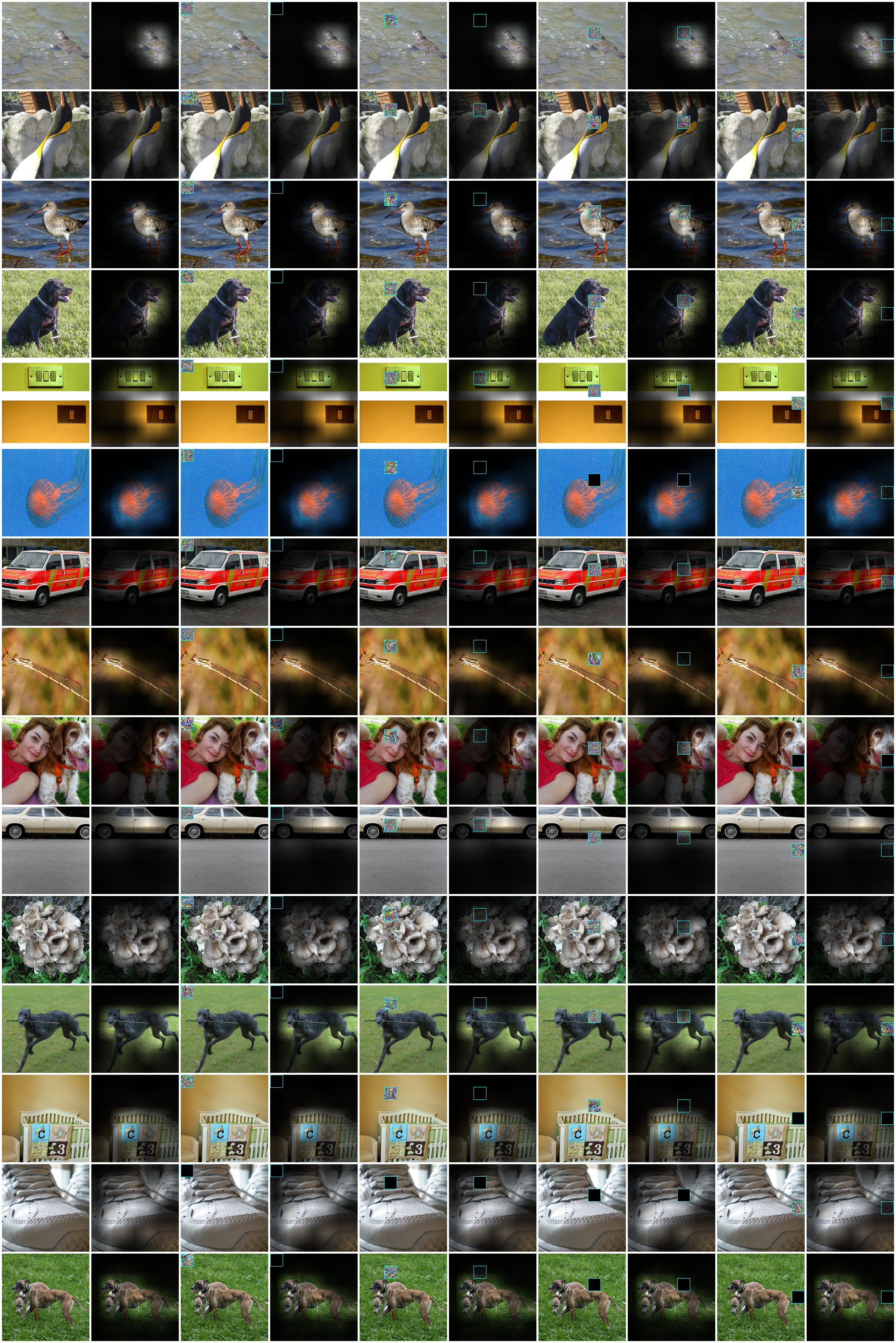}
    \caption{Averaged Feature Maps of ResNet50 as Attention with Attack Patch size of 32 on Adversarial Images}
    \label{fig:attn_res_p32}
\end{figure*}

\begin{figure*}[!ht]
    \centering
        \includegraphics[width=0.8\textwidth]{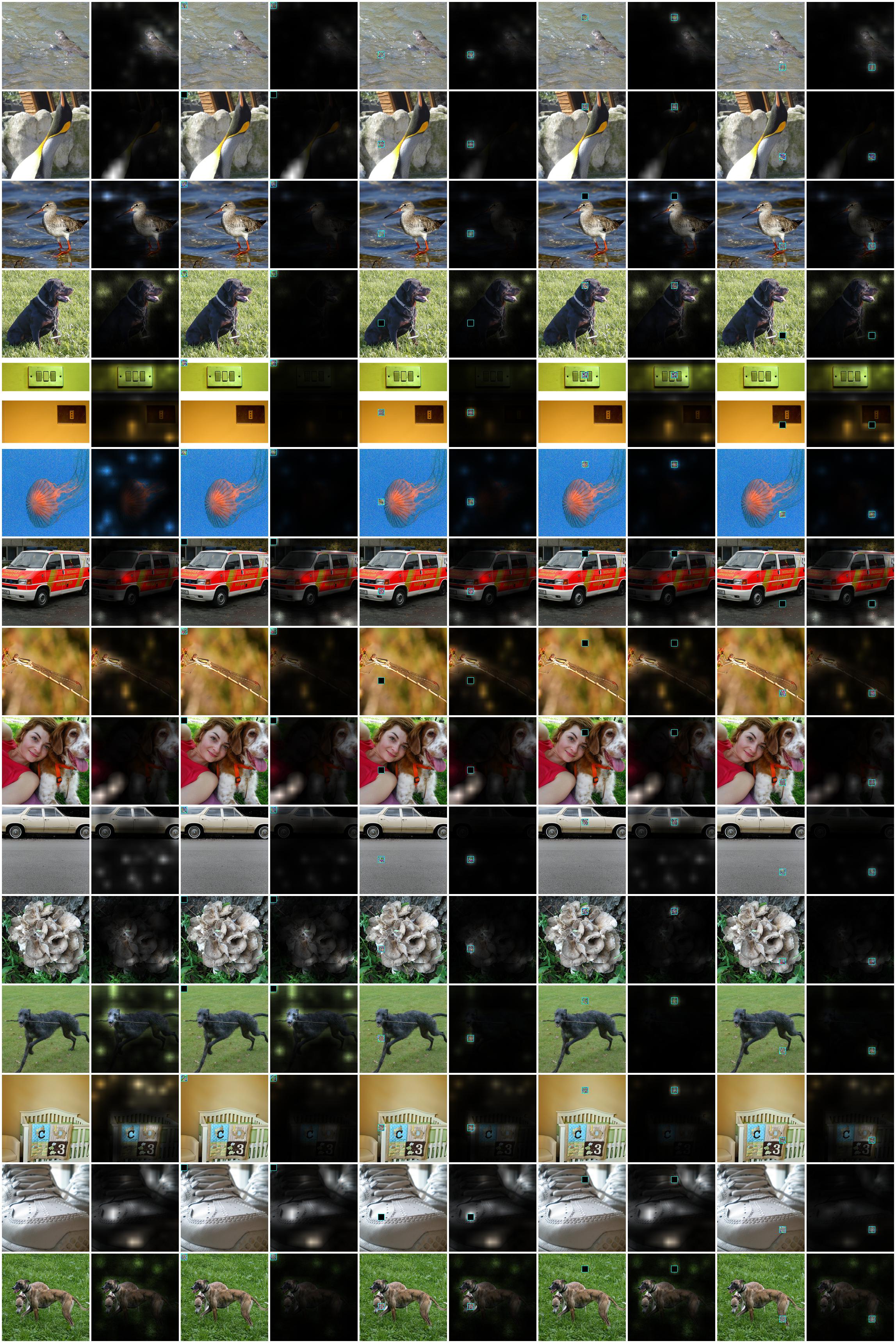}
    \caption{Rollout Attention on DeiT-small with Attack Patch size of 16 on Adversarial Images}
    \label{fig:attn_deit_p16}
\end{figure*}

\begin{figure*}[!ht]
    \centering
        \includegraphics[width=0.8\textwidth]{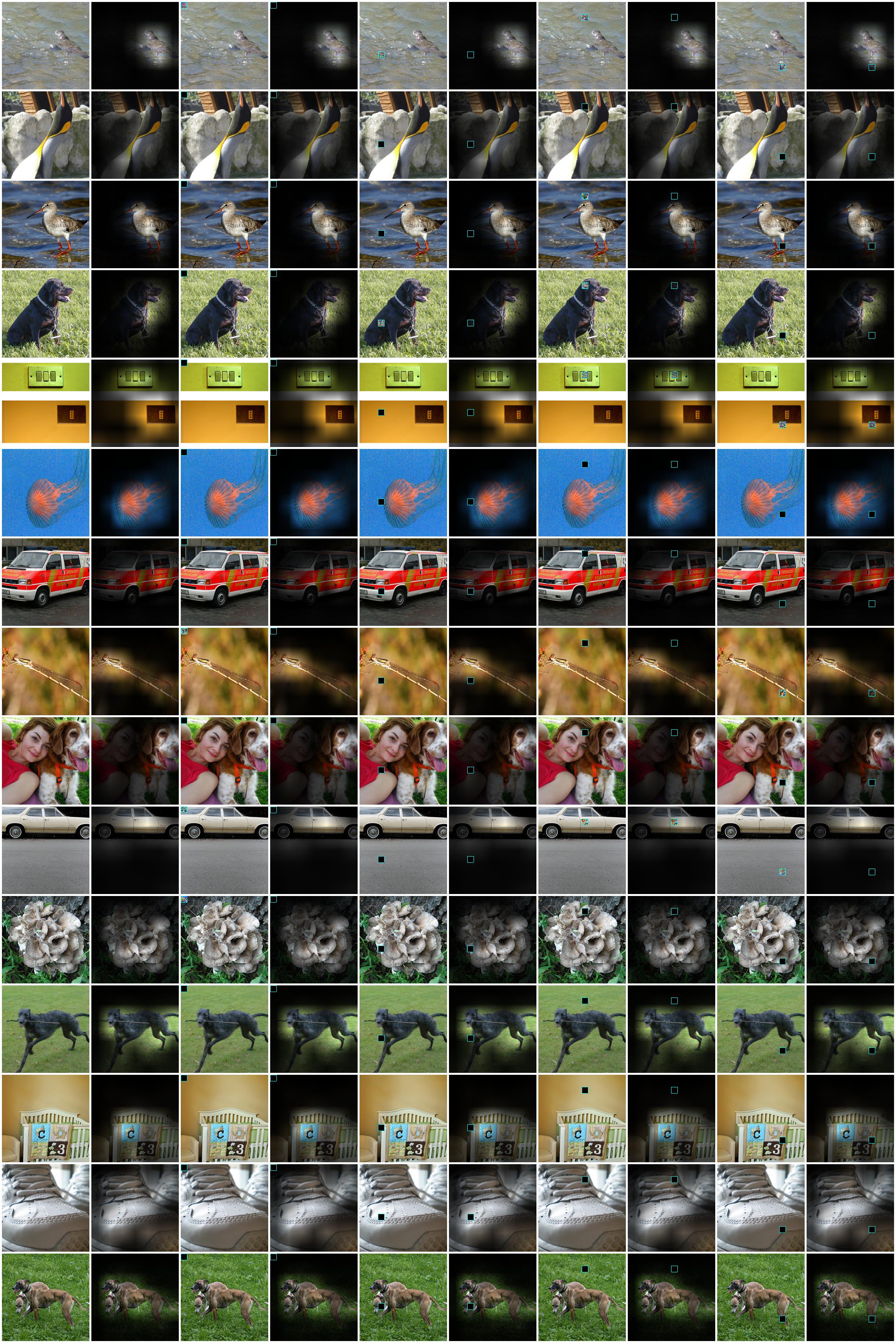}
    \caption{Averaged Feature Maps of ResNet50 as Attention with Attack Patch size of 16 on Adversarial Images}
    \label{fig:attn_res_p16}
\end{figure*}

In this appendix section, we show more Attention Rollout on DeiT and Feature Map Masks on ResNet. The adversarial patch noises with different patch sizes are shown (i.e., P=16 and P=32) in Fig.~\ref{fig:attn_deit_p32},~\ref{fig:attn_res_p32},~\ref{fig:attn_deit_p16}, and~\ref{fig:attn_res_p16}. In each row of these figures, we fist show the clean image and visualize the attention as a mask on the image. Then, we show the images with patch noises on different patch positions, and the attention masks are also shown following the correspond adversarial images.

\section{Attention under Natural Patch Corruption and Adversarial Patch Attack}
\begin{figure*}[!ht]
    \centering
        \includegraphics[width=0.8\textwidth]{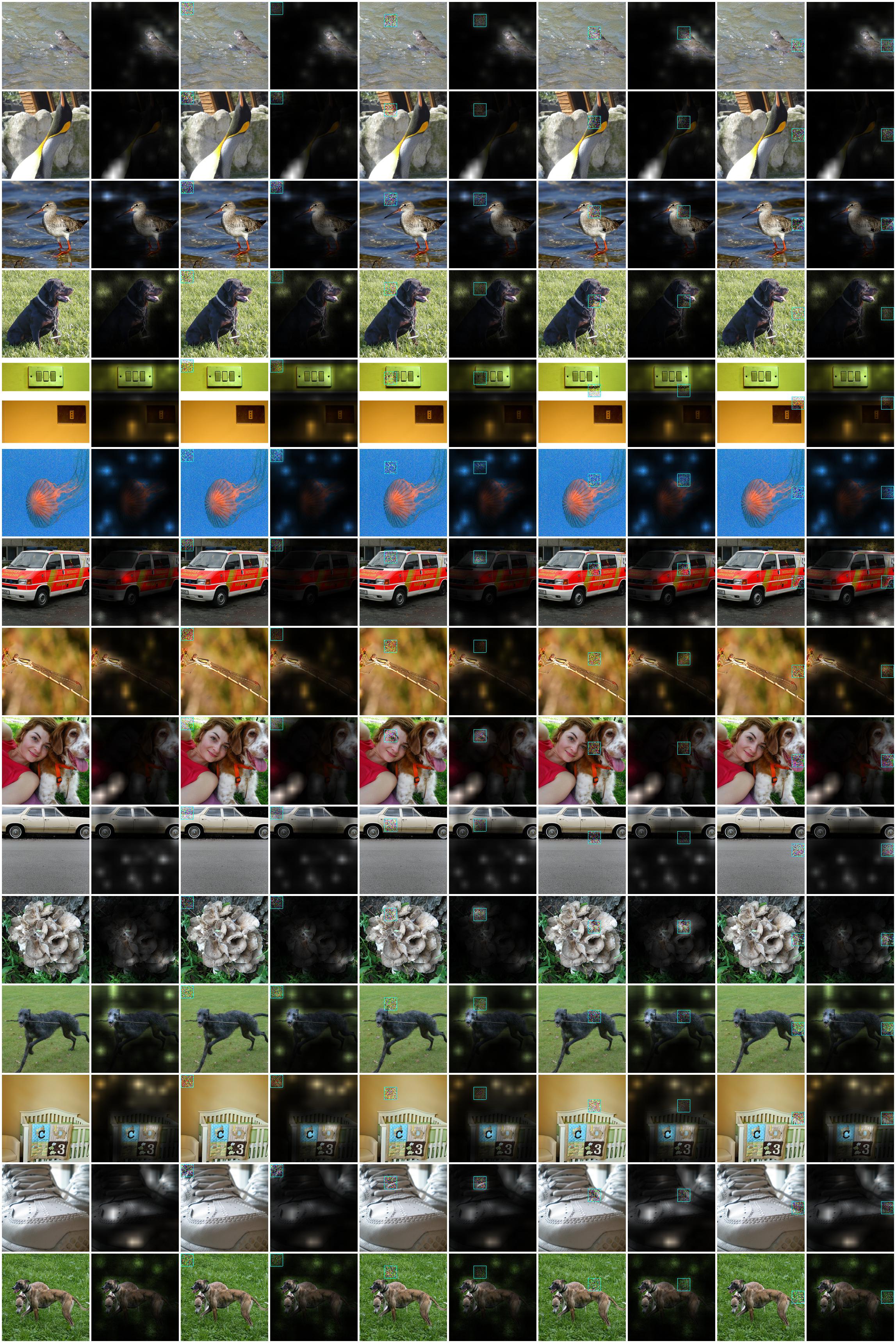}
    \caption{Rollout Attention on DeiT-small with Attack Patch size of 32 on Corrupted Images}
    \label{fig:attn_deit_p32_nat}
\end{figure*}

\begin{figure*}[!ht]
    \centering
        \includegraphics[width=0.8\textwidth]{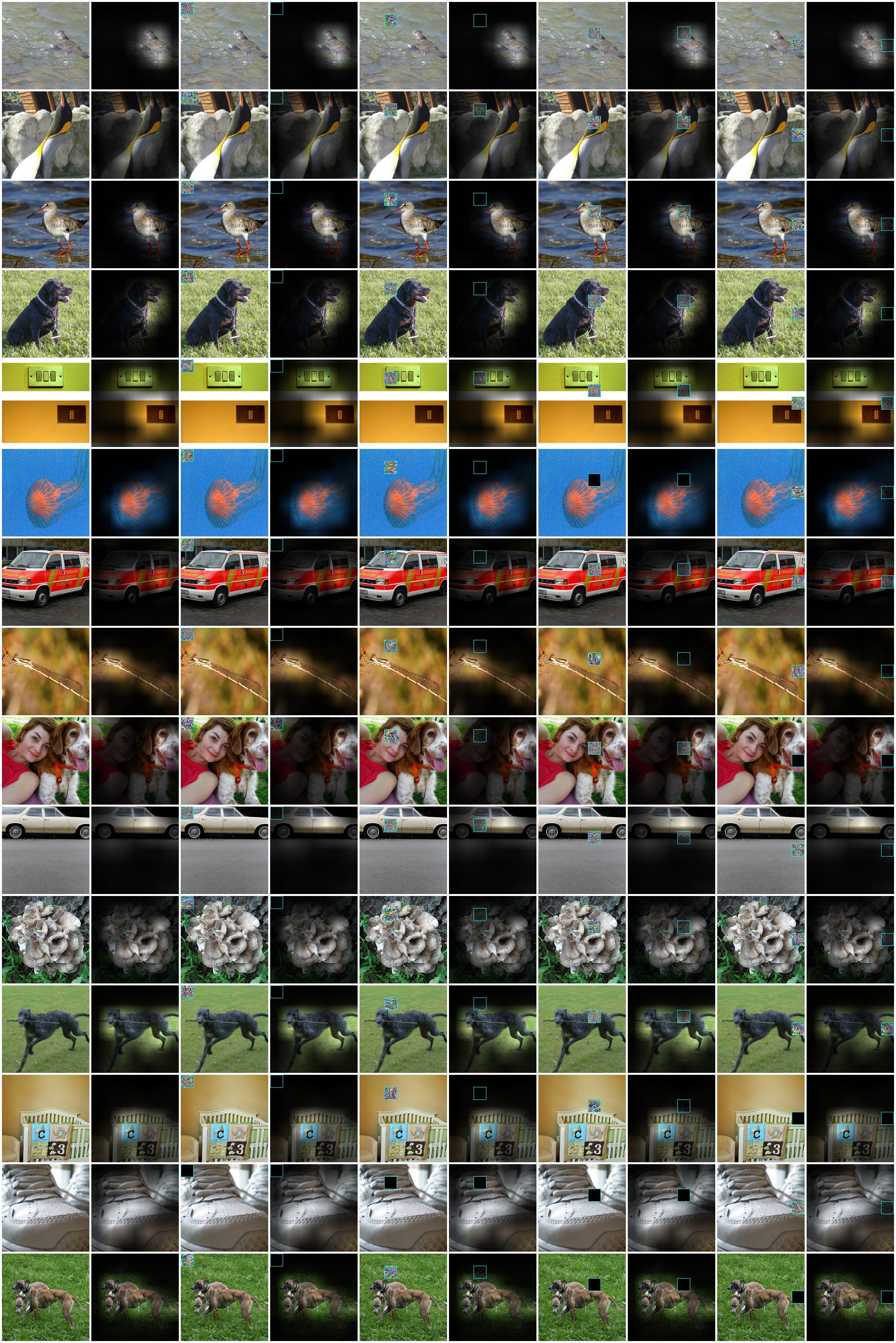}
    \caption{Averaged Feature Maps of ResNet50 as Attention with Attack Patch size of 32 on Corrupted Images}
    \label{fig:attn_res_p32_nat}
\end{figure*}

\begin{figure*}[!ht]
    \centering
        \includegraphics[width=0.8\textwidth]{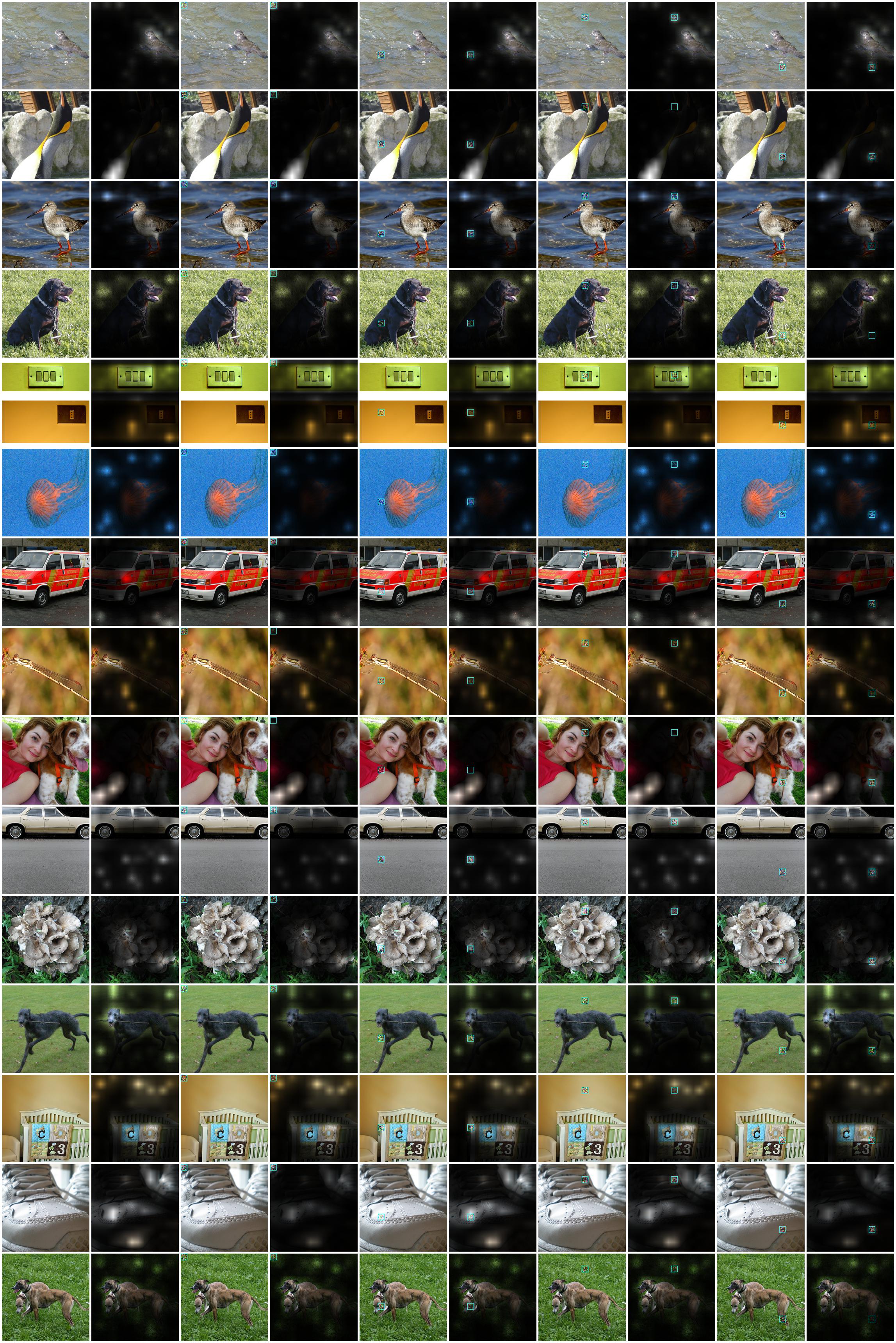}
    \caption{Rollout Attention on DeiT-small with Attack Patch size of 16 on Corrupted Images}
    \label{fig:attn_deit_p16_nat}
\end{figure*}

\begin{figure*}[!ht]
    \centering
        \includegraphics[width=0.8\textwidth]{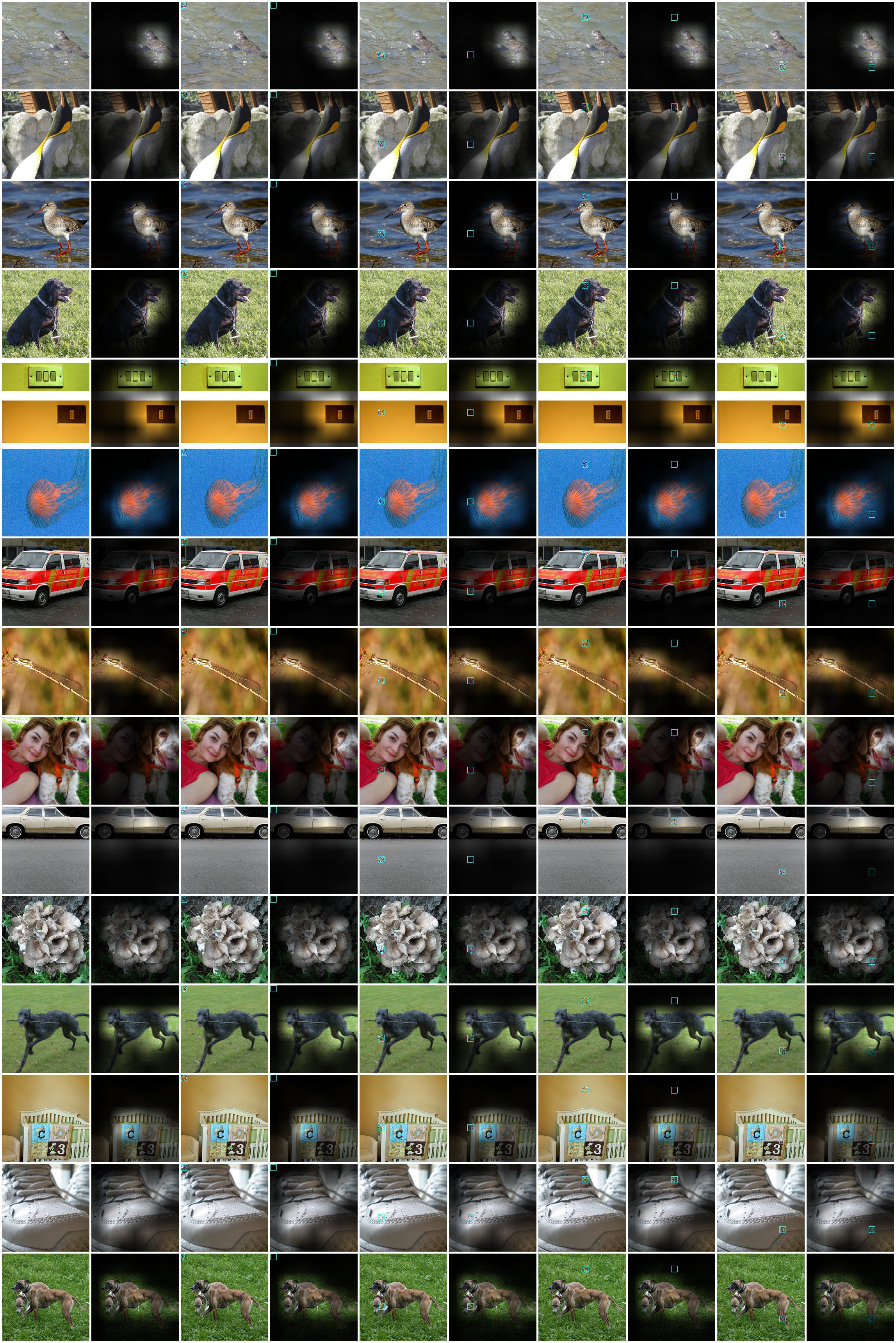}
    \caption{Averaged Feature Maps of ResNet50 as Attention with Attack Patch size of 16 on Corrupted Images}
    \label{fig:attn_res_p16_nat}
\end{figure*}

The rollout attention on DeiT and Feature Map mask on ResNet on naturally corrupted images are shown in Fig.~\ref{fig:attn_deit_p32_nat},~\ref{fig:attn_res_p32_nat},~\ref{fig:attn_deit_p16_nat}, and~\ref{fig:attn_res_p16_nat}. We can observe that ResNet treats tha corrupted patches as normal ones. On DeiT, the attention is slightly distract by naturally corrupted patches when they are in the background. However, the main attention is still on the main object of input.   

\section{Fooling Rates of Each Patch on ResNet50 and DeiT-small}
\label{app:FR_pattern_Res50_DeiTsmall}

The FRs in different patch positions of DeiT are similar, while the ones in ResNet are center-clustered. A similar pattern can also be found on DeiT-small and ResNet50 in Fig.~\ref{fig:FR_pattern_small_res50}.

\begin{figure*}[t]
    \centering
    \begin{subfigure}[b]{0.4\textwidth}
        \includegraphics[width=\textwidth]{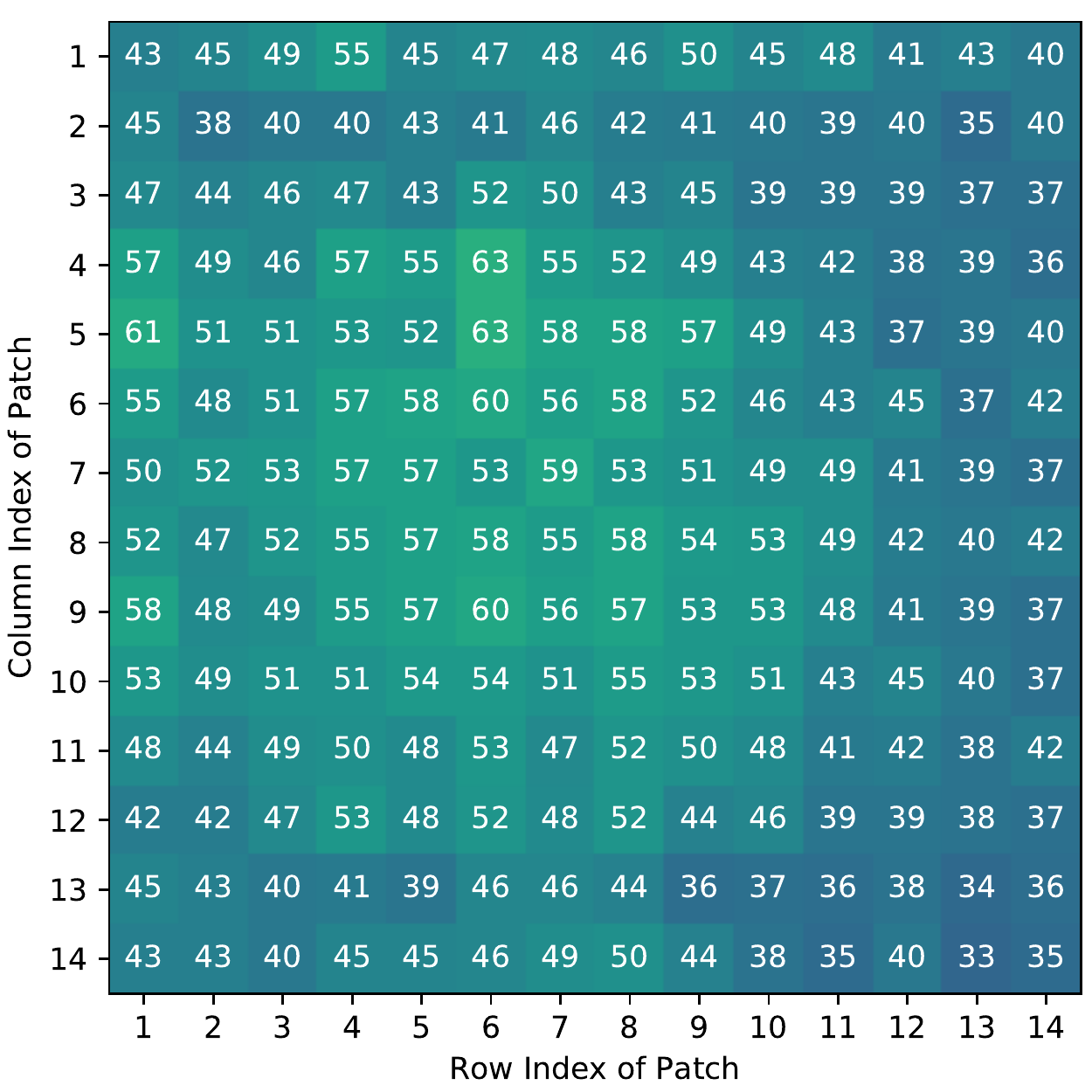}
    \caption{FRs of ResNet18 on Corner-biased Data}
    \label{subfig:corner_bias_pattern}
    \end{subfigure} \hspace{0.2cm}
    \begin{subfigure}[b]{0.4\textwidth}
        \includegraphics[width=\textwidth]{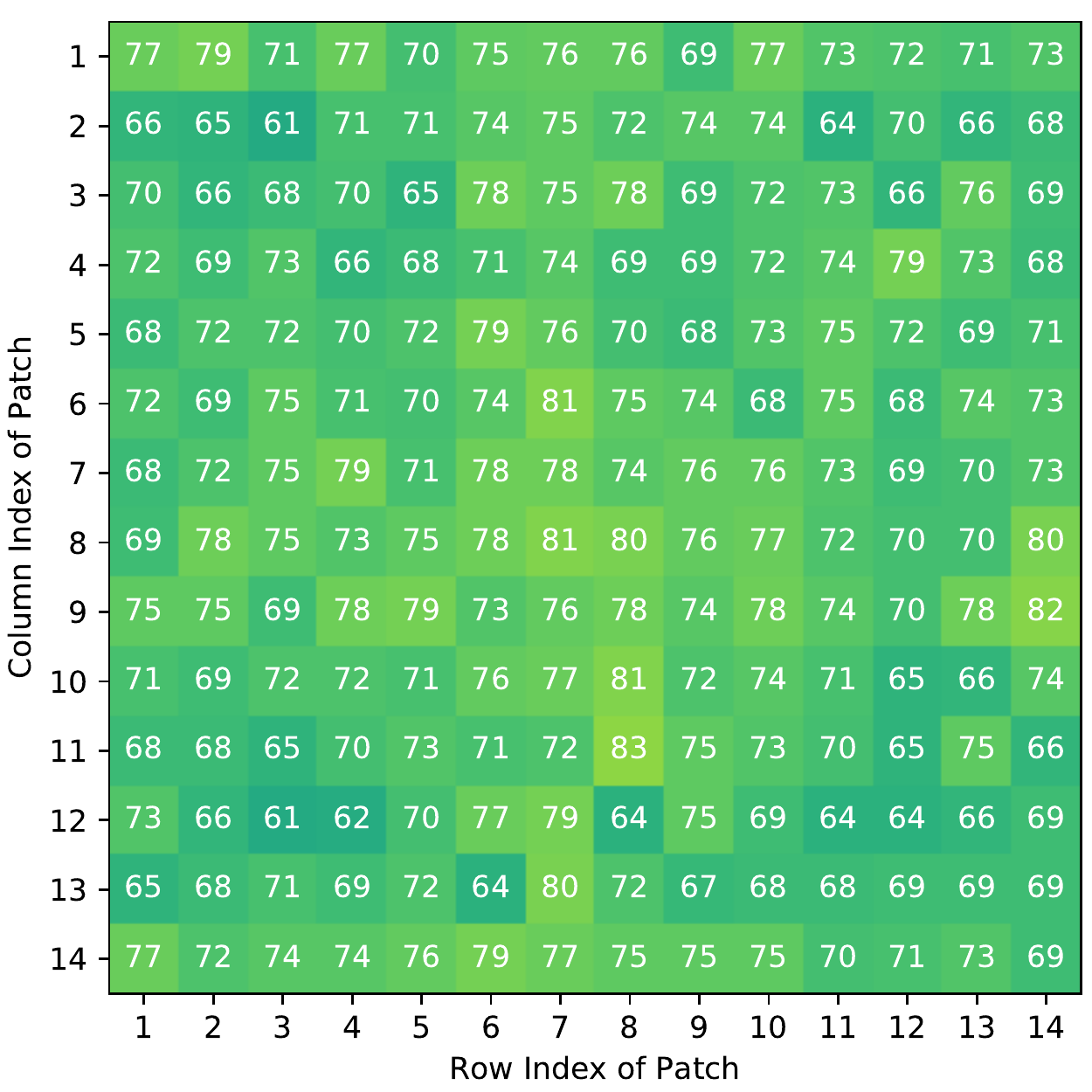}
        \caption{FRs of DeiT-tiny on Center-biased Data}
        \label{subfig:center_bias_pattern}
    \end{subfigure}
    \caption{Patch Attack FR (in \%) in each patch position is visualized on ResNet18 and DeiT-tiny on biased data.}
\end{figure*}

\section{Fooling Rates of Each Patch on ResNet and DeiT on Corner-biased Data}
\label{app:FR_pattern_biased}
In the coner-biased image set, the FR on ResNet is still center-clustered, as shown in Fig.~\ref{subfig:corner_bias_pattern}.

\section{Fooling Rates of Each Patch on ResNet and DeiT on Center-biased Data}
In the center-biased image set, the FR on DeiT is still similar on different patch postions, as shown in Fig.~\ref{subfig:center_bias_pattern}.

\section{Transferability of Adversarial Patches across Images, Models, and Patch Positions}
\label{app:transfer}

As shown in Tab.~\ref{tab:transfer_imgs}, the adversarial patch noise created on a given image hardly transfer to other images. When large patch size is applied, the patch noises on DeiT transfer slightly better than the ones on ResNet.

\begin{table}[!ht]
\caption{Transferability of adversarial patch across images}
\label{tab:transfer_imgs}
\begin{center}
\footnotesize
\setlength\tabcolsep{0.03cm}
\begin{tabular}{ccccc}
\toprule
Models & ResNet50 &  DeiT-small &  ResNet18 & DeiT-tiny \\
\midrule
across images (Patch Size=16) & 3.5  & 2.1  & 3.4 & 6.4 \\
across images (Patch Size=112) & 8.1  & 13.4  & 10.6 & 21.5 \\
\bottomrule
\end{tabular}
\end{center}
\end{table}

The transferbility of adversrial noise between Vision Transformer and ResNet has already explored in a few works. They show that the transferability between them is remarkablely low. As shown in Tab.~\ref{tab:transfer_models}, the adversarial patch noise created on a given image does not transfer to other models.

\begin{table}[!ht]
\caption{Transferability of adversarial patch across models}
\label{tab:transfer_models}
\footnotesize
\centering
\setlength\tabcolsep{0.14cm}
\begin{tabular}{c cccc}
\toprule
& \multicolumn{4}{c}{Patch Size=16} \\
Models & ResNet50 &  DeiT-small &  ResNet18 & DeiT-tiny \\
\midrule
ResNet50 & - & 0.3 & 0.16  & 2.2 \\
DeiT-small & 0.04 & - & 0.09 & 1.79 \\
\midrule
ResNet18 & 0.09 & 0.22 & -  & 1.9 \\
DeiT-tiny & 0.04 & 0.13 & 0.06 & - \\
\bottomrule
\end{tabular}

\begin{tabular}{c cccc}
\toprule
& \multicolumn{4}{c}{Patch Size=112} \\
Models & ResNet50 &  DeiT-small &  ResNet18 & DeiT-tiny \\
\midrule
ResNet50  & - & 5.25 & 8  & 11.75 \\
DeiT-small & 5.5 & - & 9.25 & 12.25 \\
\midrule
ResNet18  & 5.75 & 5 & -  & 12 \\
DeiT-tiny & 5.5 & 5 & 9.25 & -\\
\bottomrule
\end{tabular}
\end{table}

When they are transfered to another patch, the adversarial patch noises are still highly effective. However, the transferability of patch noise can be low, when the patch is not aligned with input patches. The claim on the patch noise with size of 112 is also true, as shown in Tab.~\ref{tab:transfer_posi}. 

\begin{table}[t]
\caption{Transferability of adversarial patch across patch positions}
\label{tab:transfer_posi}
\begin{center}
\footnotesize
\setlength\tabcolsep{0.14cm}
\begin{tabular}{ccccc}
\toprule
Model  & ResNet50 &  DeiT-small &  ResNet18 & DeiT-tiny \\
\midrule
across positions (0, 4) & 6.25 & 5.25 & 11.25  & 12.75 \\
across positions (0, 16) & 5.75 & \textbf{34.5} & 11.5  & \textbf{54} \\
across positions (0, 64) & 6 & 22 & 9.5  & 30.75 \\
\midrule
across positions (4, 0) & 6.5 & 5.75 & 9.75  & 12.5 \\
across positions (16, 0) & 7.25 & \textbf{35} & 10.25  & \textbf{54} \\
across positions (64, 0) & 5.5 & 18.25 & 9.25  & 31 \\
\midrule
across positions (4, 4) & 6 & 4.75 & 8.5  & 13.5 \\
across positions (16, 16) & 4.5 & \textbf{18.5} & 9  & \textbf{33} \\
across positions (64, 64) & 6 & 9.75 & 8.25  & 17.5 \\
\bottomrule
\end{tabular} \vspace{-0.3cm}
\end{center}
\end{table} 

\section{More Settings and Visualization of Adversarial Examples with Imperceptible Noise}
\label{app:vis_imper}
In the standard adversarial attack, the artificial noise can be placed anywhere in the image. In our adversarial patch attack, we conduct experiments with different patch sizes, which are multiple times the size of a single patch. The robust accuracy under different attack patch sizes is reported in Tab.~\label{tab:imper_attack}. We can observe that DeiT is more vulnerable than ResNet under imperceptible attacks.

\begin{table}[t]
\caption{Adversarial Patch Attack with Imperceptible Perturbation . FRs are reported in percentage.}
\label{tab:imper_attack}
\begin{center}
\footnotesize
\setlength\tabcolsep{0.06cm}
\begin{tabular}{c cccc}
\toprule
Model  & PatchSize=16 &   PatchSize=32 &   PatchSize=112 &  PatchSize=224 \\
\midrule
ResNet50 & 2.9 & 20.9 & 98.3 & 100 \\
DeiT-small & 4.1 & 38.7 & 100 & 100 \\
\midrule
ResNet18 & 3.1 & 26.0 & 99.1  & 100 \\
DeiT-tiny & 11.2 & 46.8 & 100 & 100 \\
\bottomrule
\end{tabular}
\end{center}
\end{table}

\begin{figure*}[!ht]
    \centering
    \begin{subfigure}[b]{0.9\textwidth}
        \includegraphics[width=\textwidth]{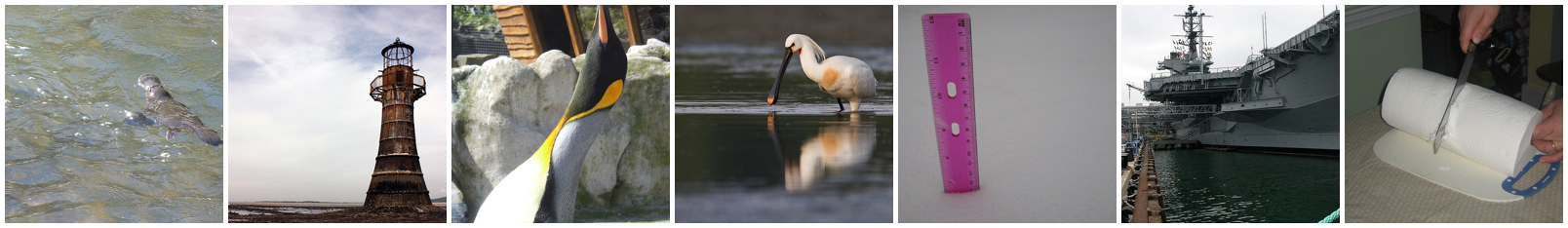}
    \caption{Clean Images}
    \end{subfigure}
    \begin{subfigure}[b]{0.9\textwidth}
        \includegraphics[width=\textwidth]{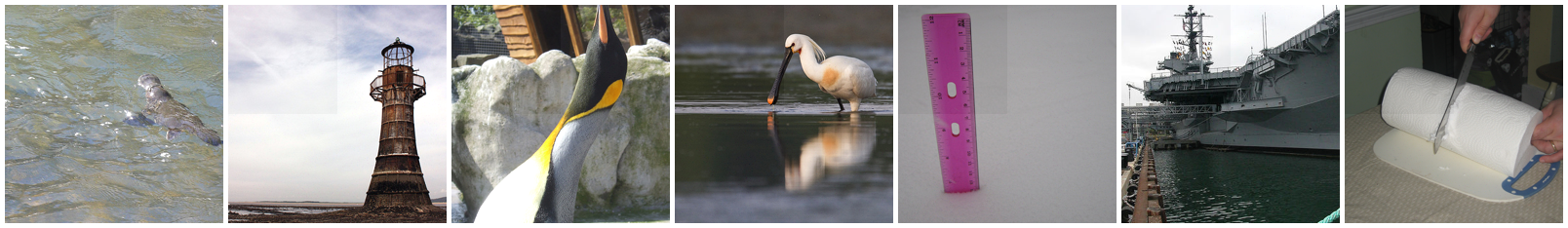}
    \caption{Adversarial Examples on ResNet18}
    \end{subfigure}
    \begin{subfigure}[b]{0.9\textwidth}
        \includegraphics[width=\textwidth]{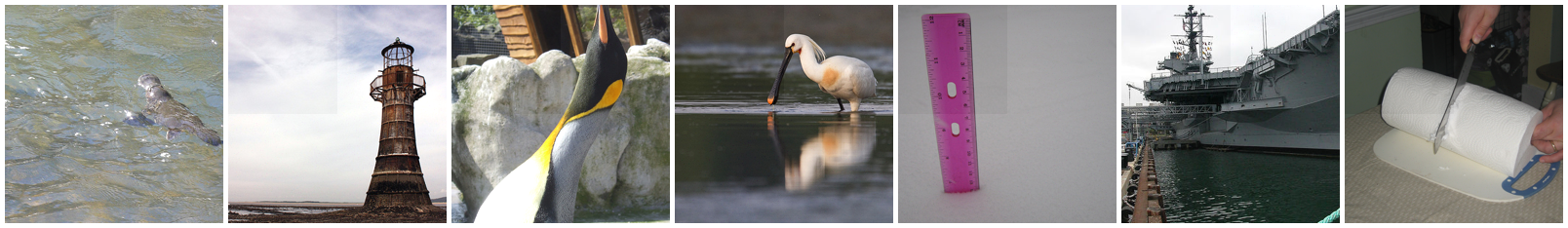}
        \caption{Adversarial Examples on DeiT-tiny}
    \end{subfigure}
    \begin{subfigure}[b]{0.9\textwidth}
        \includegraphics[width=\textwidth]{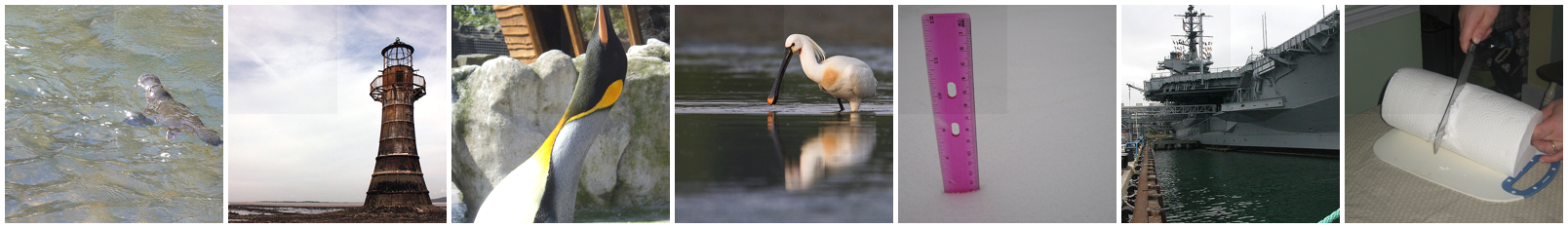}
    \caption{Adversarial Examples on ResNet50}
    \end{subfigure}
    \begin{subfigure}[b]{0.9\textwidth}
        \includegraphics[width=\textwidth]{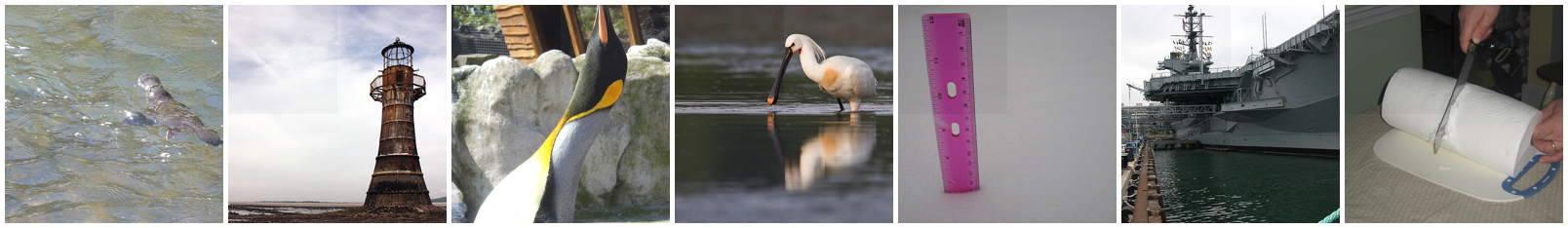}
        \caption{Adversarial Examples on DeiT-small}
    \end{subfigure}
\caption{Visualization of Adversarial Examples with Imperceptible Patch Noise: The adversarial images with patch noise of size 112 in the left-upper corner of the image are visualized. Please Zoom in to find the subtle difference.}
    \label{fig:adv_imp_vis}
\end{figure*}

The clean images and the adversarial images created on different models are shown in Fig.~\ref{fig:adv_imp_vis}. The adversarial perturbations created with imperceptible patch attack are imperceptible for human vision. 

\section{Visualization of Adversarial Patch Noise}
\label{app:vis_unbounded}
\begin{figure*}[!ht]
    \centering
    \begin{subfigure}[b]{0.48\textwidth}
        \includegraphics[width=\textwidth]{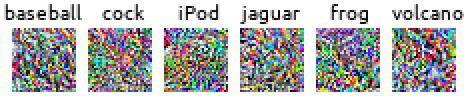}
    \caption{Patch Noise on ResNet50 under the 1st Setting}
    \label{setting1_res50}
    \end{subfigure} \hspace{0.2cm}
    \begin{subfigure}[b]{0.48\textwidth}
        \includegraphics[width=\textwidth]{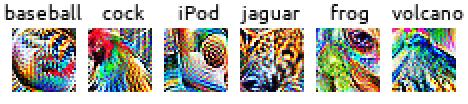}
    \caption{Patch Noise on ResNet50 under the 2nd Setting}
    \label{setting2_res50}
    \end{subfigure}
    
    \begin{subfigure}[b]{0.48\textwidth}
        \includegraphics[width=\textwidth]{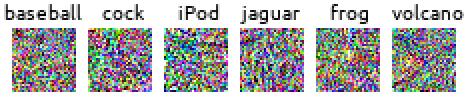}
        \caption{Patch Noise on DeiT-small under the 1st Setting}
        \label{setting1_deit_s}
    \end{subfigure} \hspace{0.2cm}
    \begin{subfigure}[b]{0.48\textwidth}
        \includegraphics[width=\textwidth]{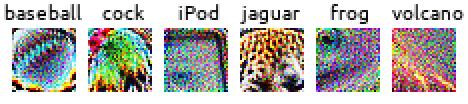}
        \caption{Patch Noise on DeiT-small under the 2nd Setting}
        \label{setting2_deit_s}
    \end{subfigure}
    \caption{Visualization of Adversarial Patch Perturbations under different Settings: In the 1st setting, the patch noise is created to fool a single classification in a given patch position. The goal in the 2nd setting to mislead the classifications of a set of images at all patch positions.}
    \label{fig:adv_noise}
\end{figure*}

Besides reporting the FRs, we also visualize the adversarial patch perturbation created on ResNet and DeiT. The adversarial patch perturbation are shown in Fig.~\ref{setting1_res50} and~\ref{setting1_deit_s}. We are not able to recognize any object in the target class. 

Following Karmon et al. 's LaVAN, we enhance the attack algorithm where we place the patch noise on different patch positions in different images in each attack iteration. From the visualization of the created noise in Fig.~\ref{setting2_res50} and~\ref{setting2_deit_s}, we can recognize the object/object parts of the target class on both ResNet and DeiT. In this section, we conclude that the recognizability of adversarial patch noise is dependent more on attack algorithms than the model architectures. 


\end{document}